\documentclass[11pt,english]{article}
\usepackage[sc]{mathpazo}
\usepackage{geometry}
\usepackage{color}
\usepackage{array}
\usepackage{bbding}
\usepackage{multirow}
\usepackage[fleqn]{amsmath}
\usepackage{amssymb}
\usepackage{amsthm}
\usepackage{graphicx}
\usepackage{natbib}
\usepackage{lscape}
\usepackage[hidelinks]{hyperref}
\usepackage{comment}
\usepackage{bm}
\usepackage[title]{appendix}
\usepackage{longtable}
\usepackage{mathtools}
\usepackage{chngcntr}
\usepackage{authblk}
\usepackage{babel}
\usepackage{dsfont}
\usepackage{subcaption}
\usepackage{threeparttable}

\makeatletter
\allowdisplaybreaks

\usepackage[mathlines]{lineno}
\usepackage{etoolbox} 

\newcommand*\linenomathpatchAMS[1]{%
	\expandafter\pretocmd\csname #1\endcsname {\linenomathAMS}{}{}%
	\expandafter\pretocmd\csname #1*\endcsname{\linenomathAMS}{}{}%
	\expandafter\apptocmd\csname end#1\endcsname {\endlinenomath}{}{}%
	\expandafter\apptocmd\csname end#1*\endcsname{\endlinenomath}{}{}%
}

\expandafter\ifx\linenomath\linenomathWithnumbers
\let\linenomathAMS\linenomathWithnumbers
\patchcmd\linenomathAMS{\advance\postdisplaypenalty\linenopenalty}{}{}{}
\else
\let\linenomathAMS\linenomathNonumbers
\fi

\linenomathpatchAMS{gather}
\linenomathpatchAMS{multline}
\linenomathpatchAMS{align}
\linenomathpatchAMS{alignat}
\linenomathpatchAMS{flalign}

\usepackage{xcolor}
\usepackage{soul}
\usepackage{algorithm}
\usepackage[noend]{algpseudocode}

\usepackage[utf8]{inputenc}
\usepackage{booktabs}

\begin{document}

\title{A multi-functional simulation platform for on-demand ride service operations}
\author[1]{Siyuan Feng}
\author[2]{Taijie Chen}
\author[2]{Yuhao Zhang}
\author[3]{Jintao Ke\footnote{Corresponding author. E-mail address: \textcolor{blue}{kejintao@hku.hk} (J. Ke).}}
\author[1]{Zhengfei Zheng}
\author[1, 4]{Hai Yang}

\affil[1]{\small\emph{Department of Civil and Environmental Engineering, The Hong Kong University of Science and Technology, Hong Kong, China}\normalsize}
\affil[2]{\small\emph{Department of Computer Science, The University of Hong Kong, Hong Kong, China}\normalsize}
\affil[3]{\small\emph{Department of Civil Engineering, The University of Hong Kong, Hong Kong, China}\normalsize}
\affil[4]{\small\emph{Intelligent Transportation Thrust, The Hong Kong University of Science and Technology (Guangzhou), Guangzhou, China}\normalsize}

\date{\today}
\maketitle

\begin{abstract}
\noindent 
On-demand ride services or ride-sourcing services, offered by transportation network companies like Uber, Lyft and Didi, have been experiencing fast development and steadily reshaping the way people travel in the past decade. Various mathematical models and optimization algorithms, including reinforcement learning approaches, have been developed in the literature to help ride-sourcing platforms design better operational strategies to achieve higher operational efficiency. However, due to cost and reliability issues (implementing an immature algorithm for real operations may result in system turbulence), it is commonly infeasible to validate these models and train/test these optimization algorithms within real-world ride sourcing platforms. Acting as a useful test bed, a simulation platform for ride-sourcing systems will thus be very important for both researchers and industrial practitioners to conduct algorithm training/testing or model validation through trails and errors. While previous studies have established a variety of simulators for their own tasks, it lacks a fair and public platform for comparing the models/algorithms proposed by different researchers. In addition, the existing simulators still face many challenges, ranging from their closeness to real environments of ride-sourcing systems, to the completeness of different tasks they can implement. To address the challenges, we propose a novel multi-functional and open-sourced simulation platform for ride-sourcing systems, which can simulate the behaviors and movements of various agents (including drivers and passengers) on a real transportation network. It provides a few accessible portals for users to train and test various optimization algorithms, especially reinforcement learning algorithms, for a variety of tasks, including on-demand matching, idle vehicle repositioning, and dynamic pricing. In addition, it can be used to test how well the theoretical models, developed in the literature for equilibrium analysis and strategic planning, approximate the simulated outcomes. Evaluated on real-world data based experiments, the simulator is demonstrated to be an efficient and effective test bed for various tasks related to on-demand ride service operations. 

\end{abstract}

{\small\emph{Keywords}: Ride-sourcing service, simulation, reinforcement learning, on-demand matching, idle vehicle repositioning}

\section{Introduction}

Recent years have witnessed a fast-growing ride-sourcing market. Transportation network companies (TNCs), such as Uber, Lyft and Didi, are utilizing smart-phone APPs to provide on-demand mobility services to passengers around the world, owing to the broad application of modern mobile communication and Global Position System (GPS). Uber, for example, is now providing different services in more than 700 metropolitan areas in 65 countries \citep{wang2019ridesourcing}. Didi, the largest ride-sharing company in China, is generating millions of daily ride-hailing demand in a single city, Beijing \citep{tong2017simpler}. In New York City, ride-sourcing vehicles are even estimated to outnumber conventional taxis 4 to 1 \citep{jiang2018ridesharing}. 

The rapid development of ride-sourcing services has raised many operational issues, including demand and supply predictions \citep{ke2021joint,feng2021multi}, dynamic pricing \citep{zha2017surge, chen2021spatial}, on-demand matching \citep{xu2018large,yang2020optimizing}, ride-pooling operations, empty vehicle repositioning \citep{zhu2021mean, lin2018efficient}, information sharing and disclosure, and rating mechanism \citep{wang2019ridesourcing}. Appropriate designs of operational strategies for ride-sourcing services require careful modelling of the market, and delicate design of optimization algorithms. At a strategical level, various mathematical models have been developed to characterize the complex but intriguing relationships between supply and demand, so as to design optimal pricing,matching and other operational strategies. At a tactic level, detailed optimization algorithms have been designed to determine the peer-to-peer matching between drivers and passengers or guide the specific idle vehicle repositioning routes, etc. 

Regarding the modelling analysis at the strategic level, researchers generally first utilize mathematical functions to depict the strategical behaviors of drivers, passengers and platforms and their endogenous interactions, and then find the supply and demand equilibrium state of the market \citep{wang2019ridesourcing}. Based on the constructed model, important influential factors for the market can be found, and managerial insights can be cast into the determination of price, wage, and matching mechanisms (e.g., regarding matching time interval and matching radius). For analytical tractability, these models generally rely on some strict assumptions on, for example, (1) equilibrium at a steady and stationary state, (2) customers' and drivers' distributions over space and time, and (3) customers' homogeneity in waiting time sensitivity and price sensitivity. It is thus unclear whether these idealized mathematical models can well describe the real-world ride-sourcing systems, which are full of stochasticity, heterogeneity and dynamics. Few efforts have been devoted to evaluating how well the outputs of these mathematical models can fit the real market outcomes, and whether the managerial insights (e.g., implications of a price increase) obtained from these models are adapted to the dynamic and stochastic environments in real-world ride-sourcing systems.  However, it is very difficult to directly implement experiments (e.g., tuning the price and wage) in real-world market, since ride-sourcing companies usually do not accept immature operational strategies to be deployed, which may results in unpredictable accidents (e.g. severe detour, excessive idling, permanent lost of some customers). Therefore, a well-performed simulator for ride-sourcing market is very important for evaluate these theoretical models and their suggested operational strategies. 

Regarding the algorithm designs at the tactic level, researchers typically formulate a specific problem (such as matching and idle vehicle repositioning) as integer programming problems for given supply and demand inputs (such as customers’ and drivers’ locations). Most of these algorithms are operated in real time. Since a real-time decision must be made within a very short time interval (e.g., 2s), these algorithms generally do not consider the strategic behaviours of customers and drivers as well as the endogenous sequential interactions among operational decisions and market status. To handle the problem, recent researchers further integrate these integer programming algorithms into a Markov decision process (MDP), which can characterize the sequential interactions between actions (decisions) and states (system dynamics). In this way, the influence of the current decision made by ride-sourcing platforms on future market states can be considered into the operational strategies for better total gain. Compared to the static integer programming algorithms which myopically focus on the decisions and objectives at the current time interval, the MDP based algorithms try to maximize the long-term rewards over a period, so that they develop more farsighted decisions. Due to the curse of dimensionality, the exact optimal solutions of these MDP based problems cannot be found; instead, approximated dynamic programming (ADP) and reinforcement learning (RL) based approaches are usually employed to find a sub-optimal solution\citep{feng2022coordinating, zhu2021mean, lin2018efficient, xu2018large}. For achieving converged results, these RL algorithms requires a training process with thousands or millions of trail and errors in an environment. Training RL algorithms in a real ride-sourcing system will incur great costs and instability to the system, because RL algorithms try different actions, some of which may be inefficient, during the training process. Therefore, a simulation platform that is close to the real system is needed to be the environment for the training and testing of RL and other optimization algorithms for ride-sourcing operations. 

Although the previous studies, especially on the tactic-level algorithm designs, have developed a variety of simulation platforms for algorithm training and testing, there are still several challenges for the design of the ride-sourcing simulator. First, early version of simulators in other studies often simplify the road network into coarse-grained grids, and the vehicles move via jump from a grid to another. Such design simplifies the simulation process, but also lose the authenticity, since the vehicles actually move on the real road network. The results generated in this way may deviate from the true ones. Second, previous simulators are typically exclusive for one or several designated operational tasks of ride-sourcing platforms. Thus, it is hard to employ them for the simulation of other types of operations, or the scenarios where multiple types of operations need to be tested. Third, the description of driver and passenger behaviors (e.g. maximal tolerable waiting time or order price for passengers, online/offline choices for drivers) are often overlooked. However, their reactions may dramatically influence the supply and demand states of the ride-sourcing systems. Fourth, researchers compare their proposed algorithms with baselines based on their own simulation platforms, which however are typically not open-sourced, making the claimed outstanding performance of the proposed algorithms less persuasive. Similar to ImageNet \citep{deng2009imagenet} for computer vision or OpenAI \citep{brockman2016openai} for general RL, an open-sourced ride-sourcing simulator is thus important for researchers on ride-sourcing service. 

To solve these challenging issues, we propose a comprehensive simulation platform for ride-sourcing service operations. The simulator can simulate the whole process of ride-sourcing platforms, including matching, driver/passenger responding, new order generation, idle vehicle repositioning, and the updation of the following market states. The simulator is completely constructed on actual road network instead of artificial grids, and can consider a variety of platform operations, such as matching, pricing and idle vehicle repositioning. The whole simulator is highly modularized, and the researchers and industrial practitioners can modify any modules in a flexible way. In the simulator, different passenger and driver behavior, such as passengers' heterogeneous tolerance waiting time in both the matching and pickup process, are sufficiently considered, in order to increase the reality. In addition, we develop interfaces within the simulator to accommodate the need of different tasks, including reinforcement learning for matching and idle vehicle repositioning, and evaluations of theoretical models. We also develop a visualization module to represent supply and demand status in different forms. Most importantly, this simulator has been open-sourced\footnote{The code is available at \url{https://github.com/HKU-Smart-Mobility-Lab/Transpotation_Simulator} while a visualization demo is available at \url{https://youtu.be/SuLIKpV2_F4}} and shared with any researchers who are interested in using the simulation platform for training/testing their algorithms and validating their theoretical models. Evaluated on a vehicle utilization based validation, some RL training tasks and one theoretical model evaluation task, we demonstrate the effectiveness of the simulation platform. In summary, this paper makes the following contributions:

\begin{itemize}
\item We develop a comprehensive, multi-functional and open-sourced simulator for ride-sourcing service, which can be used by both researchers and industrial practitioners on a variety of operational tasks. The proposed simulation platform overcomes a few challenges faced by previous simulators, including the closeness to reality, representation of customers' and drivers' heterogeneous behaviors, generalization for different tasks. 

\item The simulator provide interfaces for the training and testing of different tasks, such as testing of optimization algorithms, training/testing of reinforcement learning based approaches for matching and repositioning, evaluations of economic models for equilibrium analysis and operational strategy designs. 

\item Based on a vehicle utilization based validation task, some RL based experiments, and one task for theoretical model evaluation, the simulator is validated to be effective and efficient for ride-sourcing related researches. In the future, the simulator can be easily modified for completing other tasks, such as dynamic pricing, ride-pooling service operations, control of shared autonomous vehicles, etc. 
\end{itemize}

The remainder of this paper is organized as follows. Section 2 provides a literature review on past studies for ride-sourcing operations and some past developed simulators. Section 3 details the simulation process and specifications for the modules of the simulator, together with a validation test. Section 4 introduces some important applications of the simulator. Numerical experiments and discussions are provided in Section 5 and 6, followed by conclusions in Section 7.\par

\section{Literature Review}

\subsection{Modelling towards platform operations at strategical level}

As introduced above, there are numerous operations for ride-sourcing platform, such as matching, idle vehicle repositioning, pricing/waging. Platform managers need to design reasonable operational strategies for these operations to reach higher efficiency of the ride-sourcing platforms. Related administration also has the requirement to make regulations on the market to guarantee the social welfare and fairness for the stakeholders (e.g. platforms, passengers, drivers) of the market. To this end, researchers have put great effort into the studies of ride-sourcing market, and the researches can be mainly grouped into two streams. 

The first type of study is the theoretic modelling of the ride-sourcing market. The ride-sourcing market is first modelled with some variables and mathematical functions to depict the behavior, decision and relationship of platforms, drivers and passengers. Here, the status of the stakeholders are often represented in their equilibrium state, which shows some steady situation of the whole market. Afterwards, the functions are put together, and the solution of the corresponding equation set can be found. The variables in the functions are also be determined. Based on the solution, some interesting phenomenon can be discovered, such as the relationship within different variables (e.g., fleet size, driver wage, trip fare, driver/passenger waiting time for a ride-sourcing trip), and the managerial insights can be summarized for platform operators or administration. Such modelling methods have been applied into a variety of researches for different scenarios and operations of ride-sourcing market. \cite{yang2020optimizing} focus on the optimization of the time interval between each round of matching by solving a mathematical model in a rolling horizon procedure. In addition, they also propose a novel reward scheme (\citep{yang2020integrated}) integrated with surge pricing, where passengers pay an additional amount to a reward account on top of the regular surge price during peak hours, and use the balance in their reward account to subsidize trips during off-peak hours. They find that in some situations, passengers, drivers, and the platform will be better off under the reward scheme integrated with surge pricing.\cite{zha2018geometric} develops a model to investigate the effects of spatial pricing on ride-sourcing markets. The model is built upon a discrete time geometric matching framework that matches customers with drivers nearby. They demonstrate that a customer may be matched to a distant vehicle when demand surges, yielding an inefficient supply state. They further investigate market equilibrium under spatial pricing assuming a revenue maximizing platform, and find that the platform may resort to relatively higher price to avoid the inefficient supply state if spatial price differentiation is not allowed. More theoretic modelling based trend of researches are summarized by \cite{wang2019ridesourcing}, which is a comprehensive review for ride-sourcing related studies.

Most theoretical modelling require a well-performed simulator to complete studies. On the one hand, the theoretically discovered insights and phenomena should be validated by experiments. On the other hand, the selection of proper models under different market status is also a complicated problem, and may require experimental studies for determination.

\subsection{Algorithm designs towards real-time operations at tatic level}

In addition to theoretic modelling, another major stream for research is the algorithm design for ride-sourcing operations. Modelling based studies focus more on the extraction of macro managerial insights, while the actual practice sometime requires more dynamic real-time algorithms to improve operational efficiency. To this end, various algorithms have been designed for different ride-sourcing operations. In early studies, algorithms always optimize their goals based on the current or short-term market status. Some control theory based approaches try to solve this issue by modelling the sequential decision process of ride-sourcing operations into a Markov Decision Process, and find the exact mathematically optimal solutions under different time slot simultaneously. In this way, the short-sighted decision issue can be released. However, the solution-finding process can become extremely complicated and time-consuming, making it hard for deployment in real ride-sourcing platforms. To fix this problem, researchers adopt the idea of trial-and-error to learn the values of some important components of MDP, instead of directly determining the solution of them mathematically. Such type of approaches is called reinforcement learning, which received more and more attention from different research field, including ride-sourcing community. For example, \cite{jintao2020learning} adopt deep reinforcement learning methods to delay the bipartite matching of some orders for a potential better matching outcome (with a short pick-up time) in the incoming time intervals. \cite{lin2018efficient} propose a multi-agent reinforcement learning model that can instruct idle drivers to surrounding locations to enhance overall system efficiency over a day. The proposed multi-agent setting can consider the competition among drivers, and thus can prevent excessive accumulations of vehicles within some areas. Extensive experiments show that their proposed model remarkably outperforms the benchmark algorithms. \cite{shou2020optimal} use inverse reinforcement learning methods to decide rewards of drivers in passenger-seeking process. \cite{feng2022coordinating} utilize reinforcement learning to obtain the optimal strategies for bundled routes recommendation under the mix of ride-sourcing service and public transit system. The results show that the proposed method can improve the gains of different stakeholders simultaneously.

Careful design of ride-sourcing simulators plays an important role in the development of related algorithms. For algorithm based researches, especially reinforcement learning, some important parameters or components need to be specified via experiments (called training for reinforcement learning). Besides, the designed algorithms should also be tested in simulation environment before actual practice. As a result, a well-performed simulator is of great importance for both theoretic modelling and algorithm based ride-sourcing researches. These are all the reasons for us to develop the proposed simulator, for higher efficiency, accuracy and completeness.\par

\subsection{Simulation for ride-sourcing market}

As Introduced above, accurate and efficient simulation for ride-sourcing market is important for both modelling and algorithm related research and practice. There are a few simulators developed in previous researches. \cite{yao2020hybrid} employ data-driven multi-objective deep learning to learn ride-sourcing drivers' offline/online behavior, and integrate it into their proposed simulator. \cite{lin2018efficient} and \cite{xu2018large} both utilize a ride-sourcing simulator to help reinforcement learning algorithm training, where the vehicle movement are simplified into continuous jump within artificial grids. In \cite{zhu2021mean}, researchers propose a mean-field Markov decision process (MF-MDP) model to depict the dynamics in ride-sourcing markets with mixed agents, a representative-agent reinforcement learning algorithm to model the decision-making process of multiple drivers, together with a ride-sourcing simulator to train the algorithm. Another research \citep{ding2020simulating} propose a simulation framework to study how ride-sourcing platform should optimize pricing strategies with traffic dynamics. Cellular automaton model is utilized to simulate route choice behavior of regular vehicles and ride-sourcing vehicles in the Manhattan-like urban network. \cite{fagnant2014travel} proposes an agent-based model for shared autonomous vehicle (SAV) operations. They also explore the estimated environmental benefits via the simulator, versus conventional vehicle ownership and use. The simulation process operates by generating trips throughout a grid-based urban area, with each trip assigned an origin, destination and departure time, to mimic realistic travel profiles. \cite{nahmias2019traditional} propose an event-based modeling framework to simulate on-demand service. Behavioral models were estimated to characterize the decision-making of drivers using a GPS dataset from a major MoD fleet operator in Singapore. Other simulation platforms with various settings are developed in previous works, including \cite{nourinejad2016agent}, \cite{inturri2019multi}, \cite{yao2021ridesharing}, \cite{djavadian2017agent}, \cite{thaithatkul2019evolution}, \cite{linares2016simulation}. 

\begin{figure}[t!]
	\centering	\includegraphics[width=0.95\textwidth]{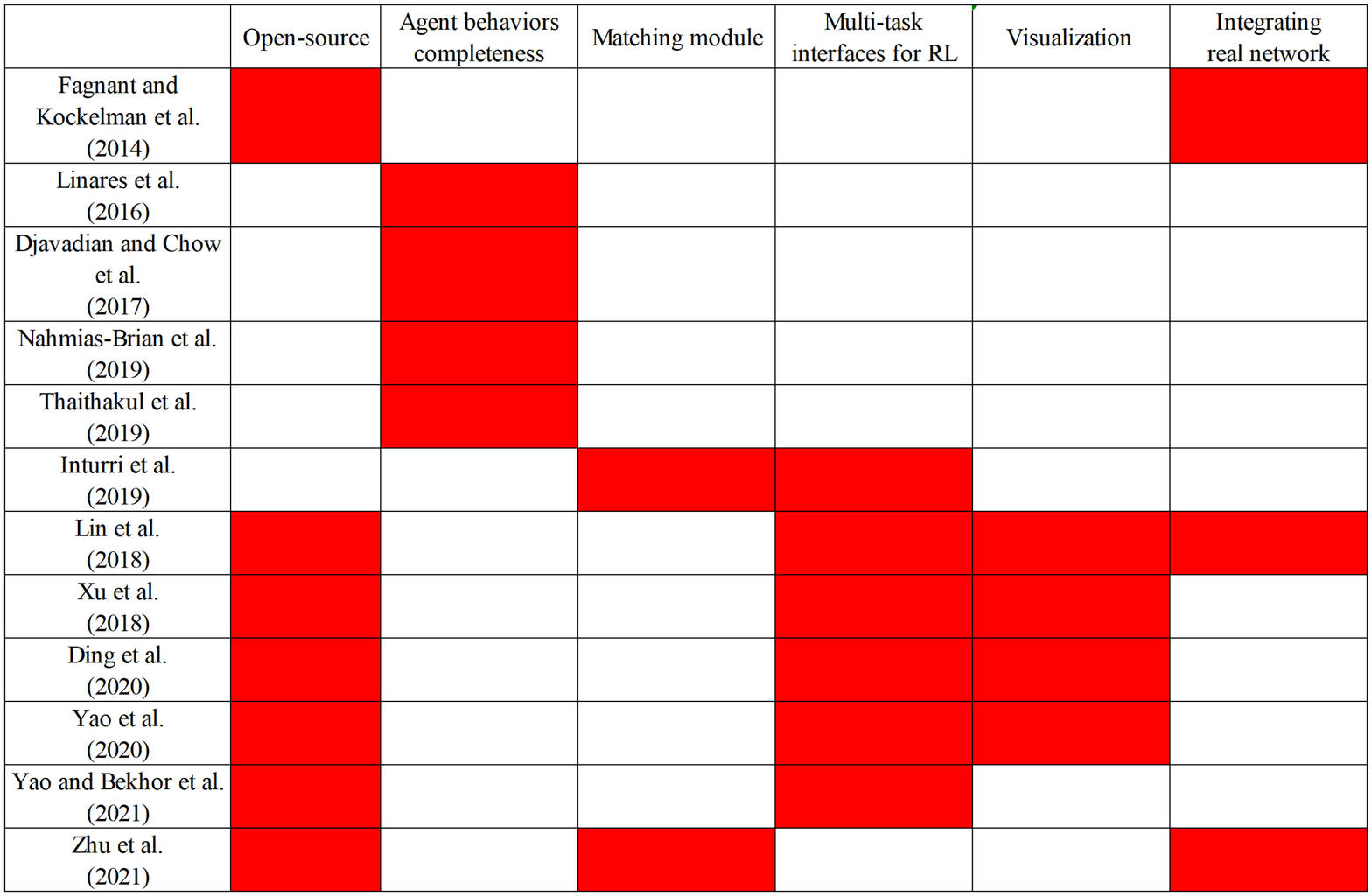}
	\caption{Components to be mainly added or improved for the previous simulators (marked by red color). }
	\label{lit table}
\end{figure}

However, all of the previous simulators have some components required to be modified and improved for better simulation authenticity, as shown in Fig. \ref{lit table}. Some of the previous simulators do not sufficiently simulate passenger and driver behavior during the simulation, such as order cancellation \citep{djavadian2017agent, nahmias2019traditional, thaithatkul2019evolution, linares2016simulation}. The accuracy of the simulation results may thus be affected. Some studies do not consider some important operations, like bipartite matching \citep{zhu2021mean, inturri2019multi}. However, such operations are actually utilized by real-world ride-sourcing platforms frequently, which should be integrated. In addition, some researches do not develop various interfaces to accommodate the requirement for different tasks, especially reinforcement learning based ones \citep{yao2020hybrid, ding2020simulating, inturri2019multi, yao2021ridesharing}. Moreover, visualization module should also be carefully designed and integrated for straightforward representation, which is overlooked by some previous studies \citep{ding2020simulating}. As mentioned above, some studies may also operate their simulators on artificial grids \citep{lin2018efficient, zhu2021mean, nourinejad2016agent, fagnant2014travel}, while higher accuracy can be reached under implementation on real road network for simulation. Also, open-source issue is an obvious yet significant problem, and most of previous simulators keep exclusive \citep{yao2020hybrid, lin2018efficient, xu2018large, yao2021ridesharing}. To overcome these issues, we develop the proposed simulator in this paper, which are specified in the following sections about how it solves the challenges above. \par 

\section{Simulation specifications}

\subsection{Simulation framework}
\begin{figure}[t!]
	\centering	\includegraphics[width=0.8\textwidth]{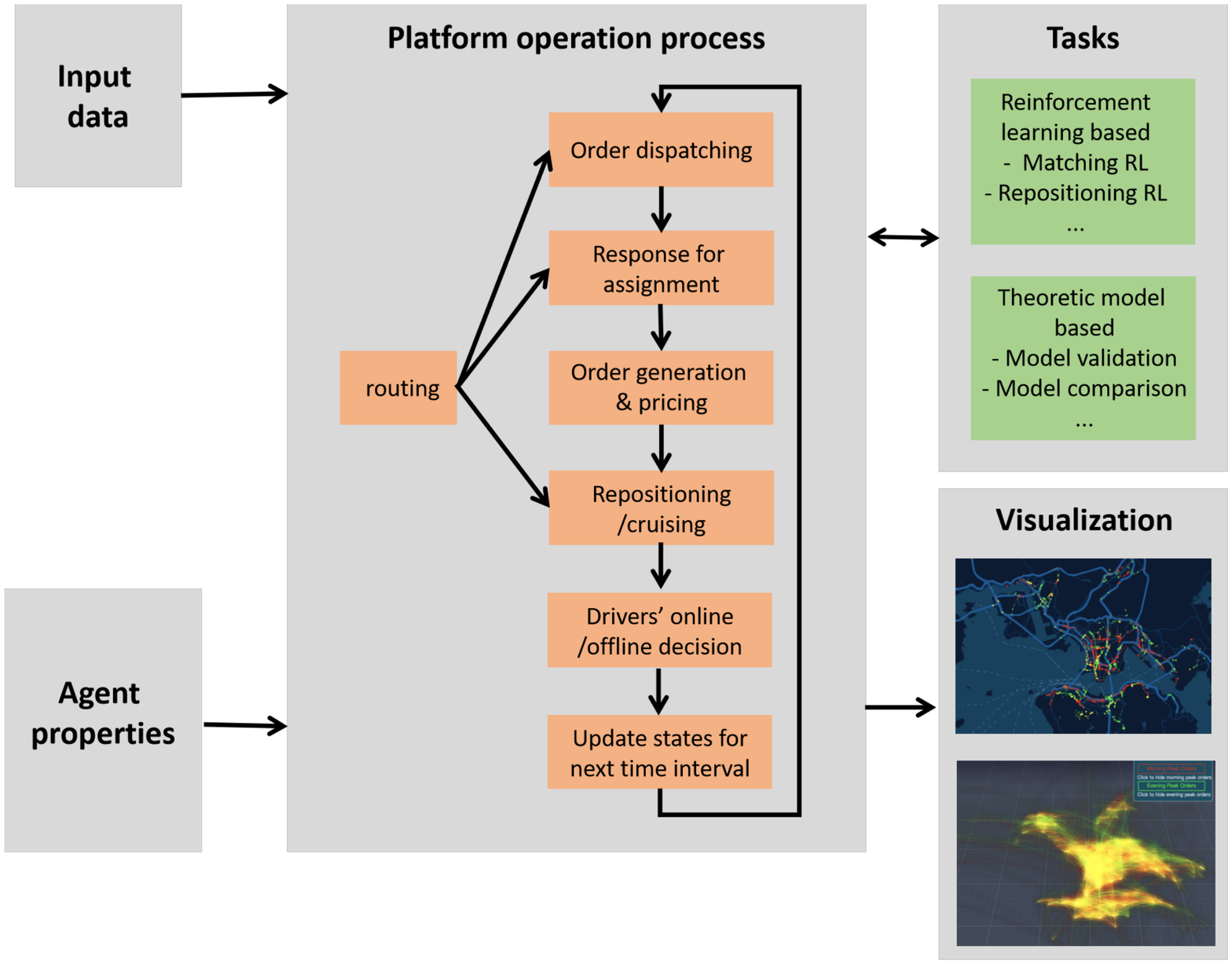}
	\caption{Simulation framework}
	\label{simulation framework}
\end{figure}

The simulation framework is represented in Fig. \ref{simulation framework}, which is comprised of five major components, including input data, agent properties, platform operation process, tasks, and visualization. The input data represents the fundamental setting of the simulation environment, under which all the other modules are implemented. Basically, these settings do not change once determined before experiments, unless there are special requirements. In addition to input data, we also define properties for passengers, drivers, and platforms within the simulation framework. These properties may change with the evolution of the state of the ride-sourcing market. The details of input data and agent propertites are specified in the next sub-section.

With the input data and agent properties given, the operation process of the platform can be constructed. During each time interval for simulation, the platform first conducts order dispatching via collecting idle vehicles and waiting passengers. Afterwards, the matching result is presented to both passengers and drivers for their reaction. In this step, some matched passengers and drivers may find the pick-up time too long, and thus abandon the matching. On the other hand, the unmatched drivers may choose to continue for waiting or drop the request, based on their maximum waiting time. Simultaneously, new orders will be generated and priced based on the given demand patterns, the current time interval, and the pricing rule by the platform. In addition, the unmatched drivers can also select idling, cruising to some other areas for searching new orders,or accept the guidance from platforms to repositioning to some given areas. In addition, the drivers may also choose dropping out of the platform based on the set patterns. Finally, all the properties for passengers, drivers and platforms are updated for the next time interval, and the whole process is repeated until the end of the simulation period. Specially, we also build a routing module to compute road node based routes given OD pairs, which is frequently utilized by other simulated operations, as shown in Fig. \ref{simulation framework}. The whole operation process introduced here will be specified in the sub-section behind.

The proposed framework provides a simulation environment to interact with different tasks for research or industry practice. As a result, the simulator is designed to flexibly accommodate the need of the tasks, such as RL for operations or theoretic model comparison. Moreover, we also develop a visualization module to represent the market status from different perspectives of demand and supply. In summary, a simulation framework is proposed to depict the whole process of the ride-sourcing market. The framework is highly complete, modularized, and user-friendly for task implementation and result representation. All the details of the simulator are open-source. For researchers, the simulator can be utilized as a shared test bed for model validation, algorithm training, testing, comparison, and demonstration. For industrial practitioners (e.g. ride-sourcing platforms and other on-demand service providers), the simulator is a free and flexible tool for algorithm pre-training for A/B test, or result representation to non-experts, such as government administration or business investors. 

\subsection{Input data and agent properties}

The input data can be divided into three groups: demand related data, supply related data and infrastructure related data. The demand related data mainly includes OD demand distribution. The demand distribution is usually extracted from historical data for the ride-sourcing market to simulate. The distribution can depict the number of people who choose ride-sourcing service during each studied time interval, as well as their OD distribution. Similarly, the supply related data is mainly about fleet distribution, including the number of vehicles emerging in the market at the start of the simulation, together with their initial locations. The final type of input data is infrastructure related ones. Currently, we mainly have road network data for this part, but it can be flexibly extended for other researches. For example, transit route data may be integrated when the simulator is modified for first-mile/last-mile issue studies. In summary, we have only three parts of input data, which reduces the difficulty for data preparation of simulation.

\begin{figure}[t!]
	\centering	\includegraphics[width=0.5\textwidth]{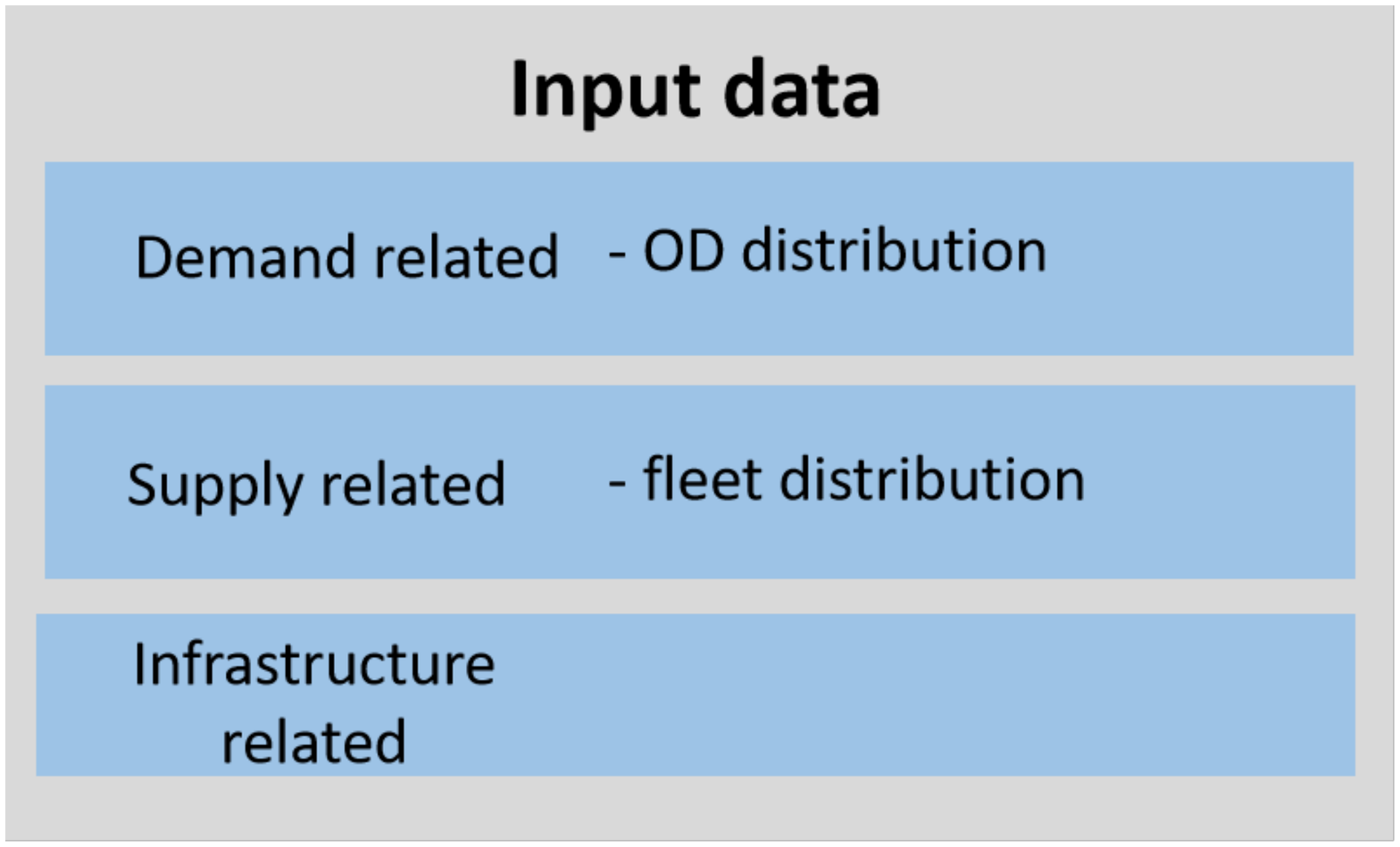}
	\caption{Input data}
	\label{input data}
\end{figure}

In addition to input data, we also define properties for passengers (demand), drivers (supply), and platforms within the simulation framework. These properties may change with the evolution of the state of the ride-sourcing market. For passengers, the related properties are order status, OD coordinates, order fare, time slot for request raising/matching/ending, pick-up time, current waiting time, maximum waiting time and cancellation probability after matching. Order status have several different values, representing unmatched, matched and waiting for pick-up, matched and under delivery, and trip finished. OD coordinates are the coordinates for origin and destination of the trip of the passenger, where the coordinate system can be defined by the simulator users. The other properties are easy to understand. Specially, the waiting time represents the time for the passengers before getting matched, and the maximum waiting time is the maximum time that a passenger can tolerate for keeping unmatched in the system. Besides, the cancellation probability after matching is utilized to depict the behavior for passenger to abandon distant matching with some vehicles. For driver side, their properties include driver status, vehicle speed, current passenger id to serve, time for idling after last trip, maximum idle time, online/offline patterns, and some itinerary related characteristics. Driver status depicts all the possible state of drivers, including cruising/repositioning, pick-up, delivery, idling (unmatched and not cruising). If the idling time of a driver become larger than the maximum idle time, the driver is considered to start searching for matching with new orders in neighboring area, and the cruising/repositioning begins. For online/offline patterns, the drivers may drop out of the system of the platform for some time, and the pattern is to depict such behavior. For itinerary related properties, there are coordinates/road node index for the driver's current location, remaining time for the current road node, list of the road nodes and their distance to each following node for the current task. These properties are added for the simulator implementation on real road network, instead of artificial grids. Finally, the platforms can control each operations via the rule for pricing, matching (including matching radius and time interval for bipartite matching) and idle vehicle repositioning. 

\begin{figure}[t!]
	\centering	\includegraphics[width=0.5\textwidth]{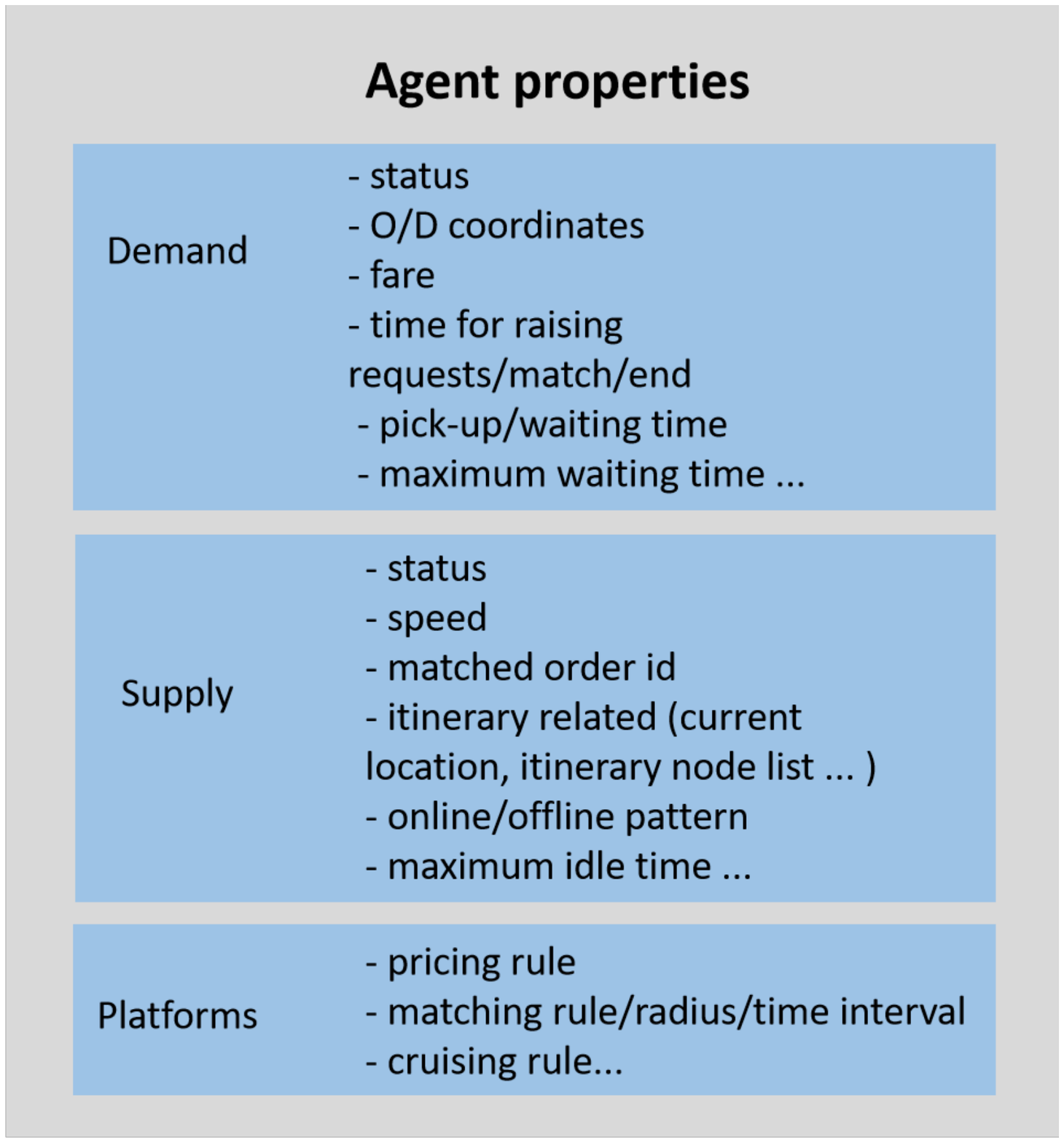}
	\caption{Agent properties}
	\label{agent properties}
\end{figure}

\subsection{Order dispatching and response after assignment}

\begin{figure}[t!]
	\centering
    \includegraphics[width=0.8\textwidth]{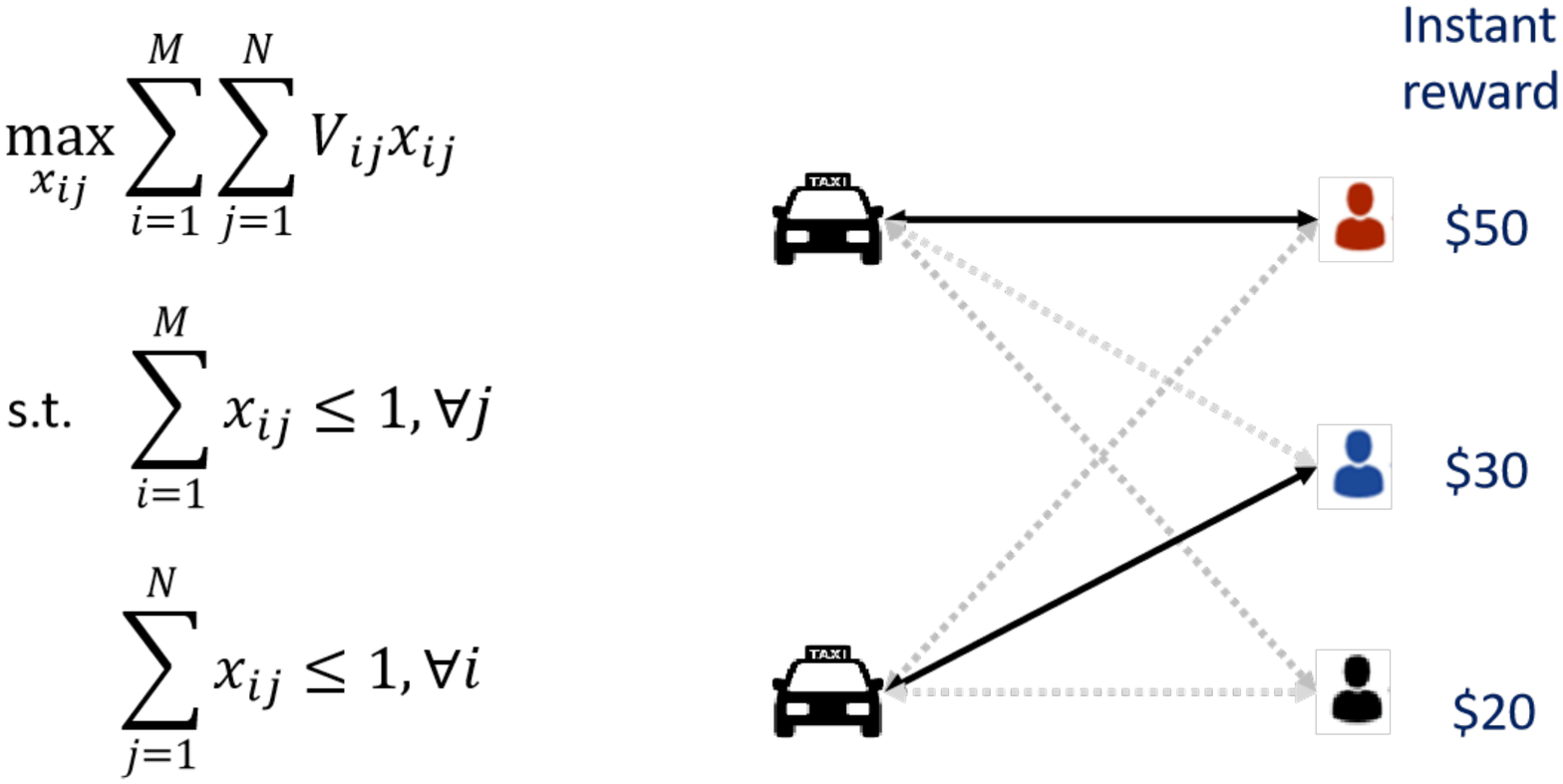}
	\caption{Bipartite matching}
	\label{bipartite matching}
\end{figure}

In the following sub-sections, we introduce the main components of platform operation process (as shown in Fig. \ref{simulation framework}) sequentially. The first one is matching operation, also called order dispatching, which is the core of ride-sourcing platforms. Here, we focus on the bipartite matching, one of the most frequently applied matching approaches for real-world ride-sourcing companies. In this process, the platforms first collect the information from waiting passengers who raised requests before, and idle vehicles at the current time interval. The passengers and vehicles are then utilized to form passenger-driver pairs together, where the selection of a pair represents the corresponding assignment of delivery. The choose of pairs for service should optimize the overall objective of ride-sourcing platforms, such as the maximization of the instant order revenue during the current time interval. This problem can be summarized into the group of bipartite matching, a classic combinatorial optimization problem. To solve the issue, the bipartite matching problem is first constructed into an Integer Linear Programming (ILP) problem, as shown in the left side of Fig. \ref{bipartite matching}. $V_{ij}$ is the service reward for driver $j$ to deliver passenger $i$. $x_{ij}$ is one if driver $j$ is finally assigned to serve passenger $i$, and otherwise zero. The first constraint requires each driver can only serve one passenger, and the second constraint requires each passenger can only be delivered by one driver. This ILP problem can be solved via some mature algorithms, such as Kuhn-Munkres (KM) methods \citep{munkres1957algorithms} or Lagrange dual decomposition based approaches \citep{feng2022coordinating}. In our simulator, all the demand and supply information can be easily extracted from the passenger and driver properties, which are continually updated. The collected information are then input into the order dispatching module, where pre-defined solution method is applied to solve the bipartite matching problem and generate the matching results. Since this module is highly modularized, it can be easily replaced with other matching rules, such as queue based ones. 

Based on the matching result, we consider different response and behaviors of passengers and drivers to increase the authenticity of simulation. For passengers, those who got matched can choose to abandon the matching by the platforms, if the pick-up distance is too large. Similarly, unmatched passengers may also abandon waiting and choose to drop out of the systems, if their cumulative waiting time is larger than the set threshold. This setting is important for ride-sourcing research, since in many theoretic studies \citep{yang2010equilibria, yang2020optimizing}, waiting time has been used for metrics to depict overall social welfare, or important internal factors that affect the euilibrium state of the ride-sourcing market. Specially, as shown in Fig. \ref{agent properties}, users can define different waiting time threshold (maximum waiting time) for different passengers. The heterogeneity of passenger behavior can thus be considered. For supply side, the matched drivers may also find the matched passengers too distant from themselves, and thus abandon the matched request. Such behavior is also carefully depicted in the proposed simulator. Moreover, the reaction of unmatched drivers are specified in the sub-section of cruising/idle vehicle repositioning.

\subsection{Order generation and pricing}

During each simulated time interval, new orders are generated based on the input OD demand pattern. The most easy way for order generation is to bootstrap from the given historical order pool, usually formed by the collection of real-world ride-sourcing data before simulation. In addition, the pricing module is integrated into the order generation module of the simulator. When new ride-sourcing orders are simulated to emerge into the market, their price are automatically given based on the pricing module, following some designed rule by users (e.g. taxi trip fare structure by Taxi and Limousine Commission for NYC data). In most studies, the pricing rule can be fixed during the simulation period. For researches involving dynamic pricing, simulator users only need to modify the pricing module directly. 

\subsection{Cruising/idle vehicle repositioning}

Generally, when the vehicle stay at the same areas and keep unmatched for a long time, the ride-sourcing driver may cruise to some other areas with possibly high demand via their experience. The cruising behavior can be very complicated, since the selection of destination for movement is influenced by multi-factors of drivers themselves, market status and physical environment \citep{wong2014modelling}. On the other hand, ride-sourcing drivers may also accept the guidance of platforms to cruise to some given destination areas. In this way, the platforms can coordinate all the relocation of idle vehicles, and balance the supply-demand gap more efficiently. Such operation is called idle vehicle repositioning, widely applied by modern ride-sourcing companies such as Didi or Uber. In the construction of the simulator, we find vehicle cruising and idle vehicle repositioning can be integrated into one module. The only difference is the one who makes the passenger-search instruction (drivers themselves for cruising, or platforms for repositioning). In this way, the simulator modules can be simplified and more easy for understanding. In the proposed simulator, the users can choose to let drivers cruise voluntarily to neighboring areas, or repositioning under instructions. The possible neighborhood for cruising/idle vehicle repositioning can be pre-defined by users of the simulator. Moreover, the cruising/repositioning rule can be flexibly adjusted within the simulator, without large interference of the implementation of other modules.

\subsection{Routing}

In many previous studies, the simulated drivers just jump from one grid to another grid of the studied area as their way of movement, where the road network is missed. However, such simplification may interfere the reliability of the testing results, since the ride-sourcing vehicles in real world are actually run on road network systems. The difference may affect the trip time given special O/D, the spatial distribution of vehicles, and thus the overall results of the simulation. To overcome the challenge, we introduce the road network system into the simulator, and the corresponding routing module to determine the path on it. We first utilize OSMnx, a Python based package to obtain geospatial data from OpenStreetMap and model the real-world road network system. For route generation, we employ the API provided by OSMnx to obtain the shortest route given some OD pair (including all the road network nodes that need to be traversed). The distance between the origin and destination can be calculated from the sum of the distances between adjacent nodes of the path. Considering there are a huge number of routes to be generated during the simulation, especially for large-scale tasks, we design a simple yet efficient approach to release the issue. Specifically, we first compute all the shortest circuit of any two road network nodes in the studied area ahead of any experiments. The computed routes are stored in some database system (e.g. Mongodb), with the corresponding OD pair as their indexes. During simulation, the simulator can directly obtain the routes by providing OD to the database system, without the need to compute routes repetitively. In this way, the simulation time can be greatly reduced, making the proposed simulator more practical.

\begin{figure}[t!]
	\centering
	\includegraphics[width=0.8\textwidth]{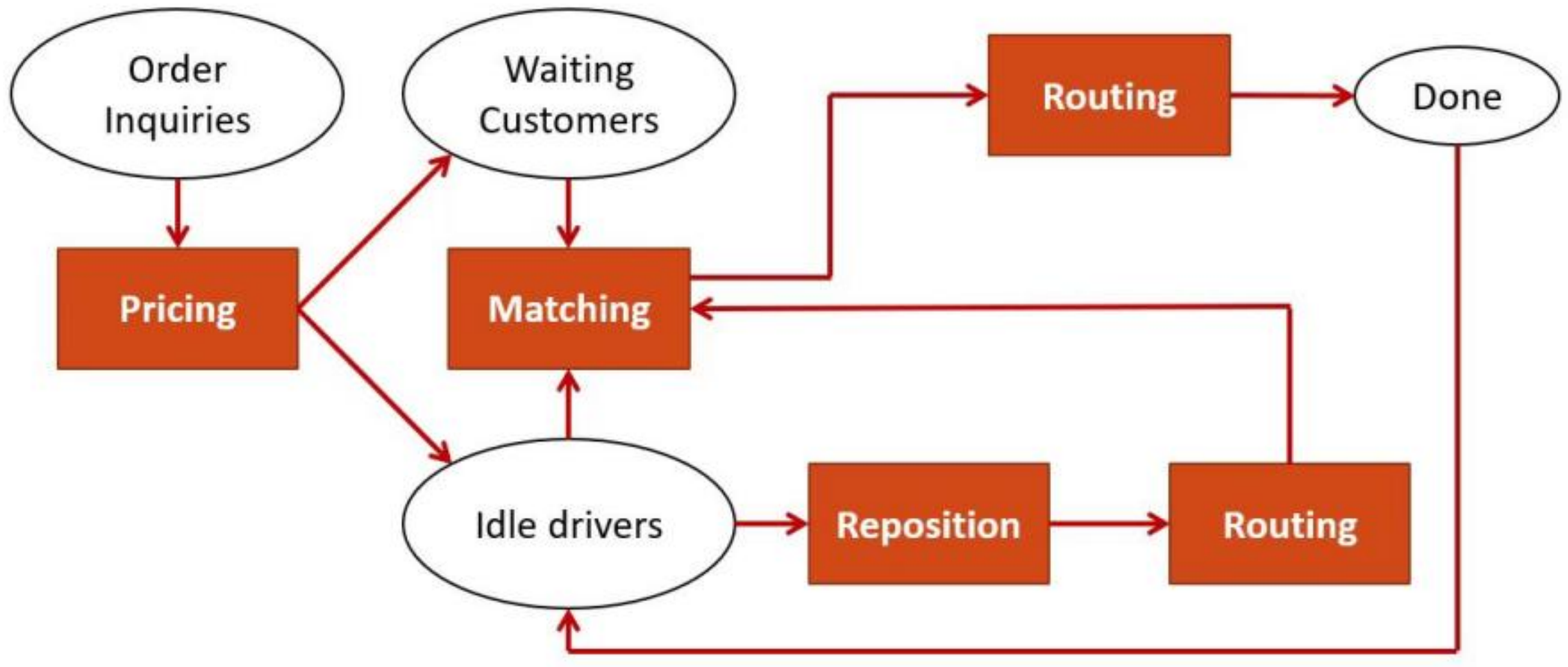}
	\caption{Utilization of routing module}
	\label{routing module}
\end{figure}

As shown in Fig. \ref{routing module}, the developed routing module is utilized in different operations within the simulation framework. In matching operation, once the matching result is provided, the routing module is utilized to generate the routes of pick-up and delivery. In cruising/repositioning module, the routing module is used for the route planning of cruising or repositioning vehicles. Since all these two operations are frequently used in the simulation, the design in our routing module can assist in releasing the computation burden, and improving the speed for the whole simulation.

\subsection{Drivers' online/offline decision and market status updation}

Unoccupied drivers in idling or cruising/repositioning state may choose to drop out of the simulated platform for relaxation or changing to other platform. In the proposed simulation, such behavior can be considered by setting the offline time for each driver before simulation. Once the time is up and the driver is not in service for some passengers, the driver will be poped out of the simulation. 

With all the previous steps for the operation process finished, the agent properties are updated for the next time interval of the simulation to simulate the evolution of market status. For passengers, their status are changed if the waiting/delivery is finished at the next time interval. Also, the waiting time will increase correspondingly for the unmatched passengers. For drivers, the status is updated also, together with their itinerary related properties. If the current task for the driver is ongoing at the next time interval, the itinerary node list keeps the same, while the driver's road node is updated to the next one. Otherwise, the whole itinerary node list will be updated based on the new task started at the next time interval. With such updation repeated, the natural transformation of the whole ride-sourcing market can be simulated.

\subsection{Visualization module}

\begin{figure}[t!]
	\centering	\includegraphics[width=0.95\textwidth]{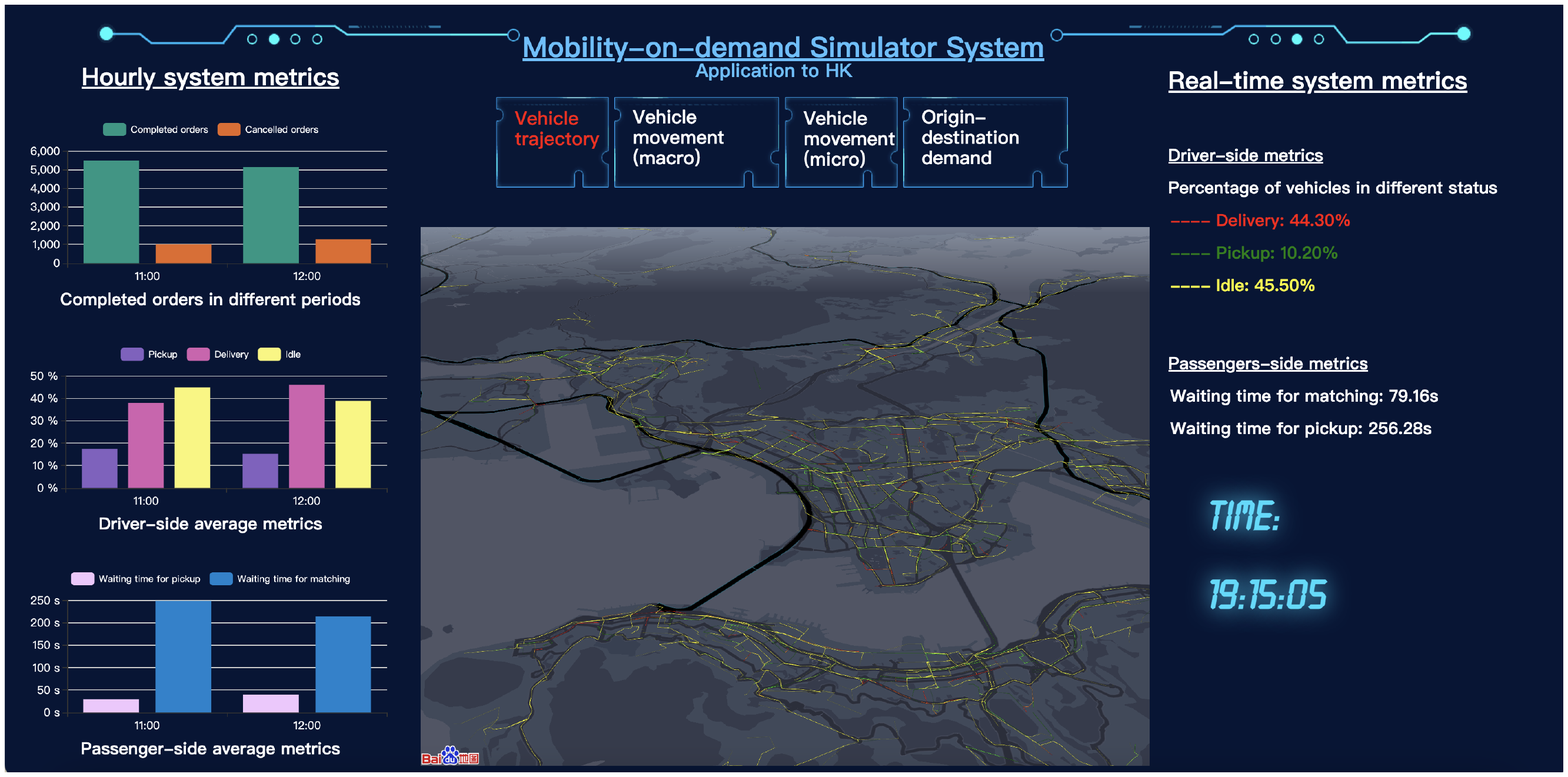}
	\caption{Visualization module for the proposed simulator}
	\label{visualization interface}
\end{figure}

The visualization module consists of three parts in its interface. As shown in Fig. \ref{visualization interface}, Left part is hourly system metrics, this section mainly shows various indicators of different time periods, including the number of completed orders, the number of order cancellations, the average pick-up, drop-off, cruise time of the driver, and the average waiting time of the passengers and the matching time.

\begin{figure}[t!]
	\centering
	\includegraphics[width=0.6\textwidth]{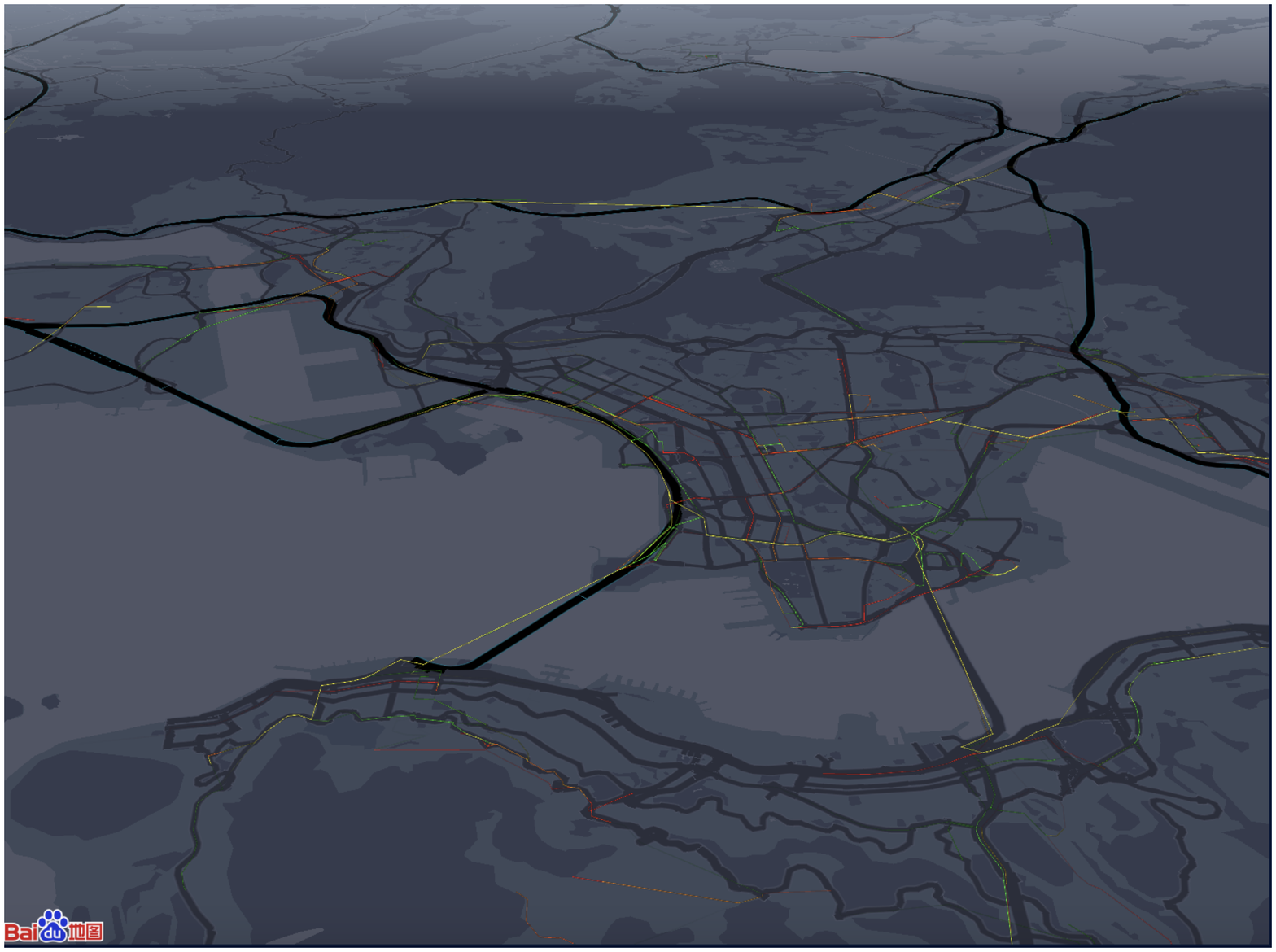}
	\caption{Trajectory map }
	\label{trajectory}
\end{figure}

\begin{figure}[t!]
	\centering
	\includegraphics[width=0.6\textwidth]{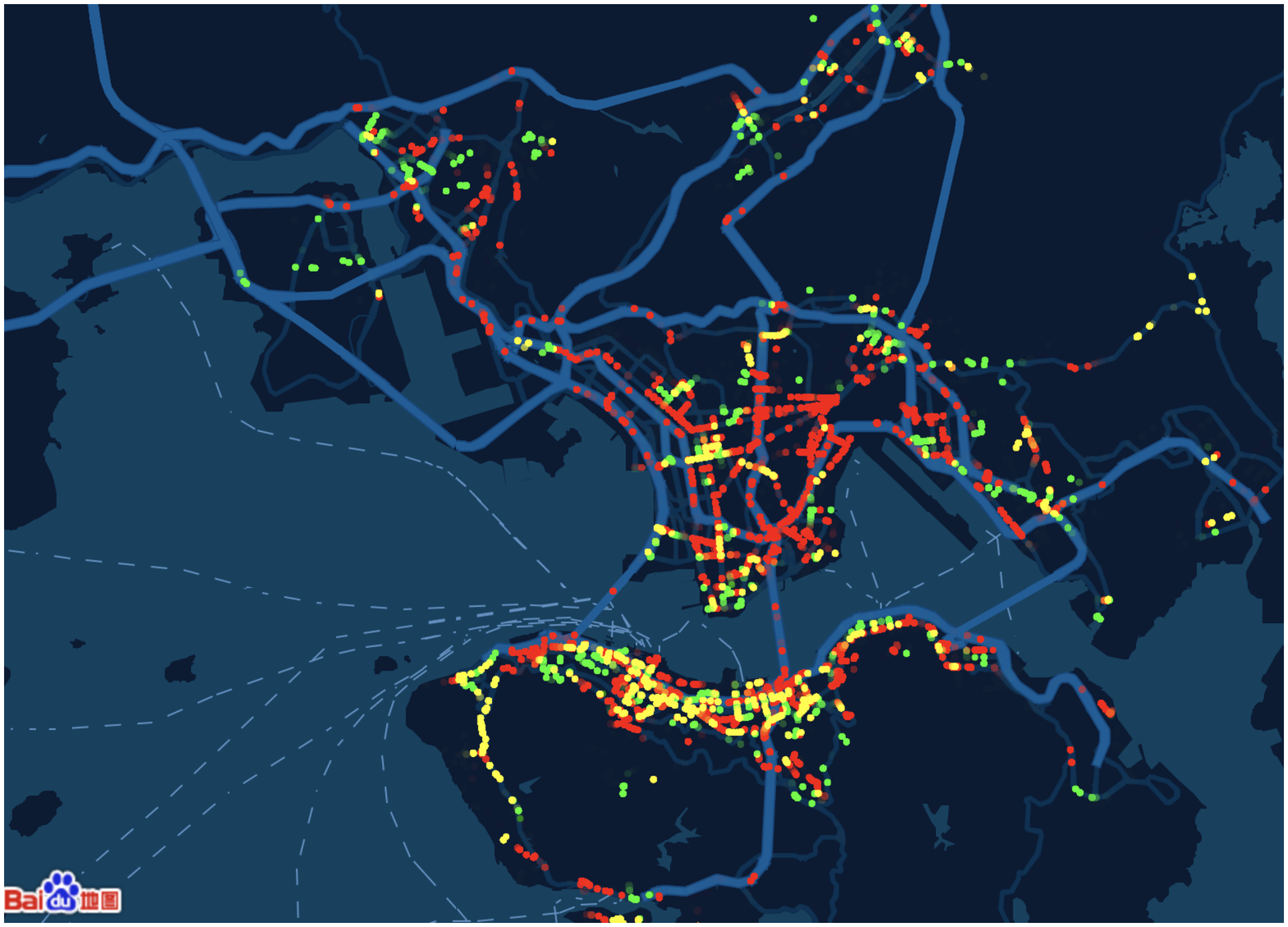}
	\caption{Vehicle movement (macro) }
	\label{vehicle movement macro}
\end{figure}

\begin{figure}[t!]
	\centering
	\includegraphics[width=0.6\textwidth]{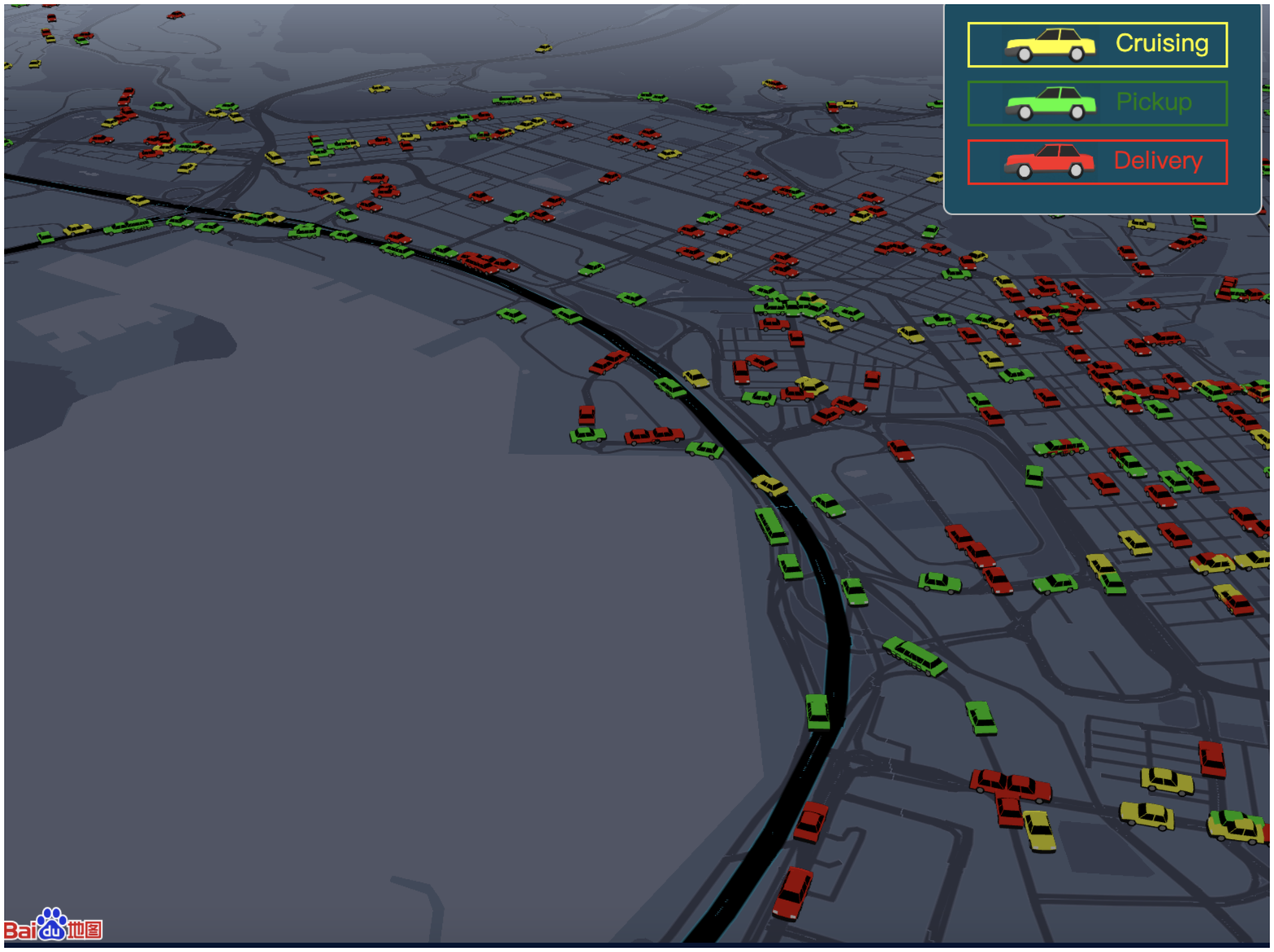}
	\caption{Vehicle movement (micro) }
	\label{vehicle movement micro}
\end{figure}

\begin{figure}[t!]
	\centering
	\includegraphics[width=0.6\textwidth]{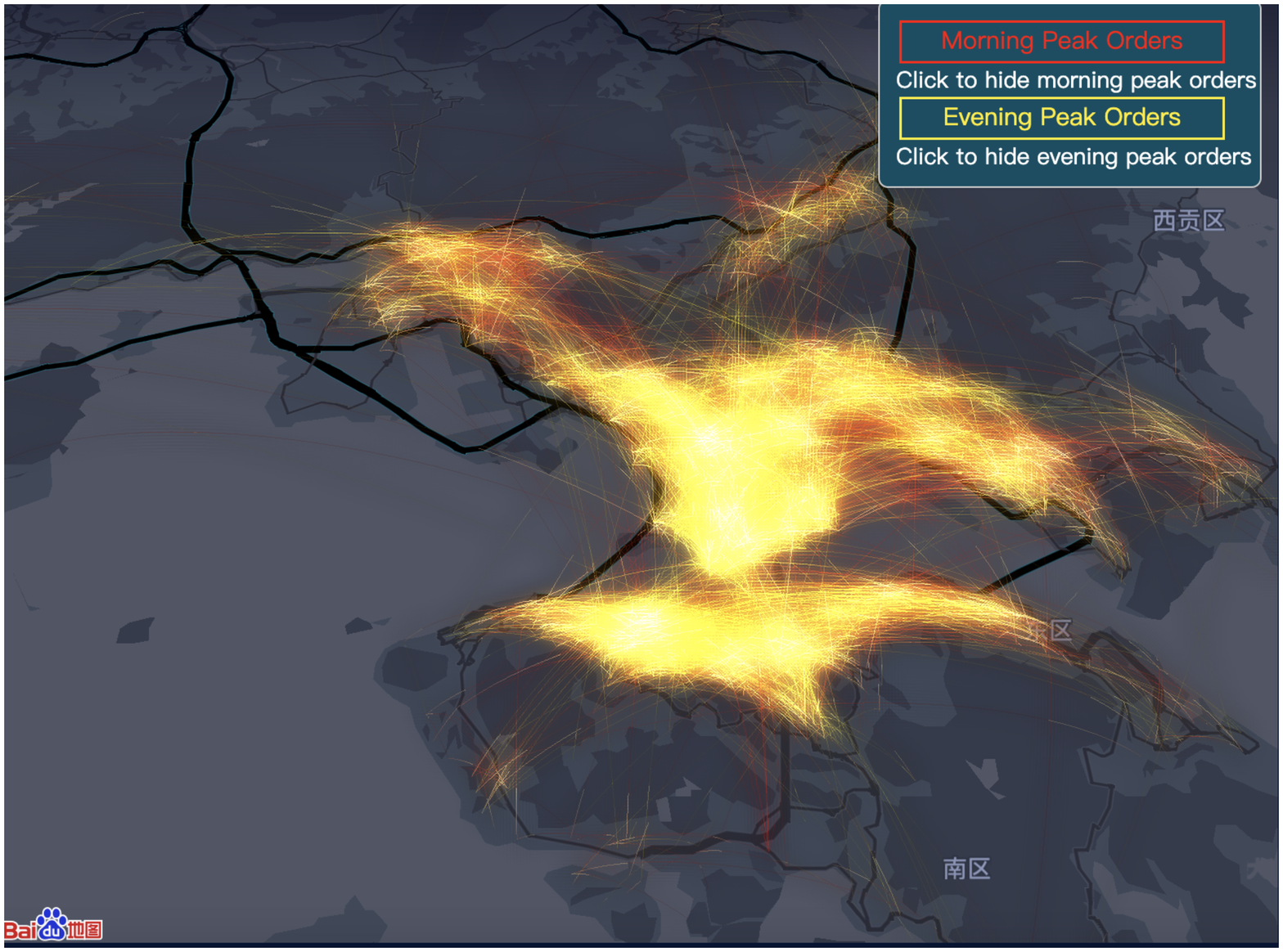}
	\caption{Origin-destination demand}
	\label{OD demand}
\end{figure}

Middle part is the main window of the visualization. Four different maps can be selected by users, including trajectory map, vehicle movement (macro) map, vehicle movement (micro) map and origin-destination demand map respectively. For trajectory map in Fig. \ref{trajectory}, the green lines indicate drivers’ trajectories on the way to pick up passengers, the yellow lines indicate drivers’ searching and cruising routes, while the red lines refer to drivers’ movements for delivering passengers from their origins to their destinations. All trajectories are located on the real transportation network. For vehicle movement (macro) in Fig. \ref{vehicle movement macro}, the green, yellow, and red dots track the real-time locations and movements of ride-sourcing vehicles in the states of picking up, searching for, and delivering passengers, respectively. For vehicle movement (micro) in Fig. \ref{vehicle movement micro}, vehicles in different colors present vehicles’ movements on the road network. In Fig. \ref{OD demand}, origin-destination demand map help to show the spatio-temporal distributions of passengers’ orders. 

Finally, the right part in Fig. \ref{visualization interface} is real-time system metrics. The included indicator is similar to the ones in left part, except that they are collected and presented in a real-time way. All the indicators and maps can dynamically transform with the simulated market status, which is a graphic presentation of the invisible back-end of the simulator.

\subsection{Validations of the simulation platform}

The modules of the simulation process has been specified in the previous sub-sections. In this part, we conduct an experiment based on the proposed simulator, in order to further validate its capability of approximating real-world ride-sourcing market. 

We select and process a taxi dataset of Hong Kong for the experiment, with complete information of drivers and passengers in the market, such as order details and the continuous records of drivers' tracks. Based on the historical information, some metrics can be formulated to evaluate the accuracy of the proposed simulation. Here, we employ the average utilization rate of vehicles as our major metrics for the following reasons: 1) it represents the average occupancy of the vehicles, which is the core of the ride-sourcing platforms and markets; 2) it is relatively insensitive to the detailed setting of the market, which helps to simplify the calibration process of the simulator. 

With the metrics determined, the utilization rate is first calculated based on the real track records, and then generated via the simulation. By comparing the difference between real and simulated utilization rate, we can demonstrate how well the proposed simulation framework approximate the reality. For some specifications in the simulation, the average vehicle speed is set to 23km/h, and the maximal pick up distance is 5km. The fleet size and spatial distribution is also calibrated based on the real taxi data. For cruising rules, when drivers are not matched, we set them to randomly cruise to nearby areas with larger historical demand to mimic the drivers' experience for order searching. In addition, we utilize $ \frac{|U_r - U_s|}{U_r}$ as the error between real and simulated utilization rate of vehicles, which are respectively represented by $U_r$ and $U_s$. 

The results are provided in the figure below, where the red curve is the utilization rate for real data, and the blue one is the utilization rate generated by the calibrated simulator. As shown in the graph, the two curves basically well matched with each other, while the error is only around 0.17. The result validates that the proposed simulator can effectively fit the real-world data after calibration, which is the fundamental for the following applications in the next section.

\begin{figure}
\centering
\includegraphics[width=0.9\linewidth]{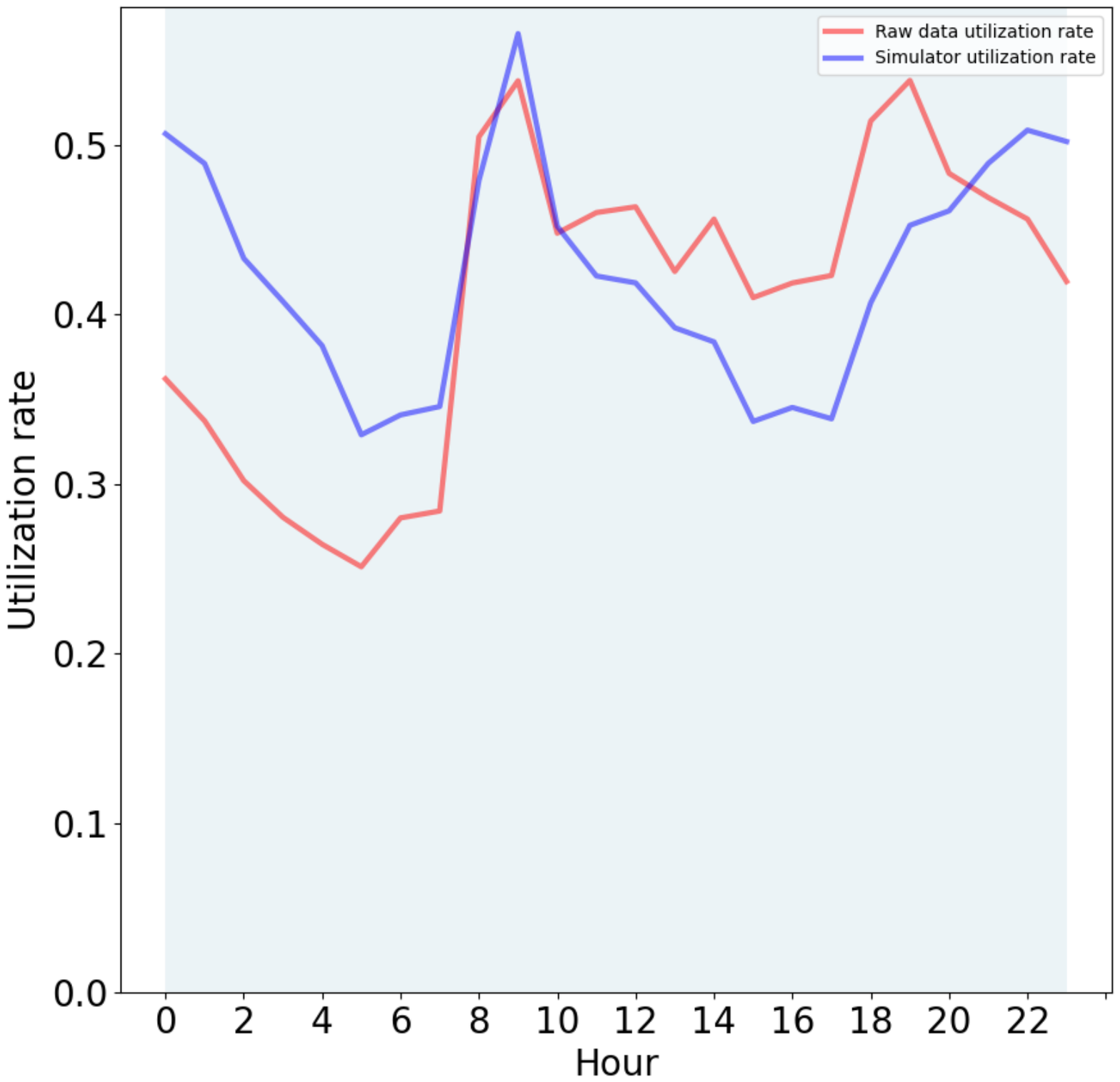}
\caption{Utilization rate}
\label{utilization rate}
\end{figure}

\section{Applications of the simulation platform}

\subsection{Theoretical model evaluation}

Researchers have been always attempted to describe the ride-sourcing markets through mathematical models. Take matching operation for example, previous studies have developed a variety of matching functions for ride-sourcing markets, including perfect matching function, Cobb-Douglas type matching function, queuing models, and some physical models. Most of these functions have demand and supply related parameters as their input, and the output is the value to represent the number of successful match. However, less is known about the applicability and performance of these matching functions, that is, under what situations each of these matching functions well characterizes the real market. Similar problems can also emerge for other mathematical functions to depict different operations. To solve these problems, researchers have to calibrate, validate, and compare the prevailing functions in the literature, and ascertain the conditions of their applicability. Since it is basically inapplicable to directly test the functions in real-world ride-sourcing platforms, such process heavily depends on numerical experiments, which are implemented on the ride-sourcing simulator. 

Usually, the whole process for such studies is specified below. Scenarios of the ride-sourcing market under different combinations of supply and demand should be first determined and input to the proposed simulator. Some key performance metrics, such as the matching rate in the market, passengers’ average matching time, passengers’ average pick-up time, and passengers’ average total waiting time, are utilized to test and compare the theoretic functions under various market scenarios. Thanks to the rich setting of agent properties of the proposed simulator, these metrics can be easily and flexibly constructed according to the research requirements. Afterwards, the true values of the key market metrics are recorded after the simulation reaches a stationary state. Although large number of repetitive experiments are required, the tricks for acceleration in the proposed simulator (e.g. introduced setting in routing module) can help to reduce the computation time while keeping simulation reliability. Finally, the estimated values of the key market metrics of each market scenario is calculated on the basis of different theoretic functions. The functions are evaluated by comparing the true value and estimated value of the key market metrics under each market.

The proposed simulator can not only assist in the implementation of the process above for aggregate model, but it also accommodates the needs of theoretic studies for disaggregate models. Aggregate models take macro and aggregate parameters as input, while disaggregate models focus on depicting the more detailed situations for different areas and individuals. In the proposed simulator, different agents can possess differentiated properties. In addition, the simulator is implemented on road network system, where the whole studied area can be easily divided into road-segment level for disaggregate models. 

\subsection{Reinforcement learning for ride-sourcing operations}

As introduced above, proposed algorithms for ride-sourcing service often require adjustment and testing for performance improvement before formal application, which is heavily dependent on the simulator. Here, we focus on the reinforcement learning task, an important emerging group of algorithms for the modern ride-sourcing platforms. Conventional algorithms usually only decide their operation strategies based on the current market status. However, the current decision made may also have impact on the future market status, which should be considered. In reinforcement learning, the decision made on different time steps is depicted into Markov Decision Process (MDP). The optimal strategies of the MDP usually consider both current and future gain, and thus can release the previous raised problem. RL based algorithms are actually developed to find the optimal solutions of MDP.

For a more specific introduction, we may take matching operation, idle vehicle repositioning and pricing as example. Usually, there are five components of a MDP, which are specified as follows for different operations.

\vspace{1.5ex}
\textbf{a. MDP for matching:}

\textbf{Agent}: Available drivers waiting for match are the agents we consider.

\textbf{State}: The state ($S$) has two components: location ($g$) and time interval ($t$). The first represents the location of the driver, and the second represents the current time.

\textbf{Action}: There are two groups of action ($A$): order serving and idling.

\textbf{Reward}: The reward ($R$) should correspond to the objective for the system. For most researches and platforms, order revenue is set to the reward.

\textbf{State transition}: State transition probability ($P_{ss^{'}}^a$) describes the probability of transfer from the current state $s$ to a certain next state $s^{'}$, after conducting action $a$. In the experiment, the state transition probability is determined by the simulator. 

\vspace{1.5ex}
\textbf{b. MDP for idle vehicle repositioning:}

\textbf{Agent}: Available drivers waiting for repositioning instruction of platforms.

\textbf{State}: The state includes location ($g$) and time interval ($t$), which is similar to that of matching operation.

\textbf{Action}: The action space is constructed by the possible destinations for idle vehicle repositioning given the current location of the considered driver. For simplicity, an auxiliary grid system can be established on the road network, and the possible destinations for repositioning become the adjacent grids of the current location. 

\textbf{Reward}: Still take the maximization of platform revenue as the overall goal. Assume a driver makes an repositioning decision at time $t$ from grid $g_i$ to $g_j$, and the expected arrival time is $t+1$. Then the reward can be set to the average order revenue for all the drivers arriving $g_j$ at $t+1$. Such design of rewards can help avoiding greedy actions that deliver excessive drivers to the area with high value of orders. 

\textbf{State transition}: Similar to the situation in matching operation, the state transition probability ($P_{ss^{'}}^a$) here represents how the drivers transform from the current state to another given the made decisions.

\vspace{1.5ex}
\textbf{c. MDP for pricing.}

The overall price for a trip is usually composed of the starting price and the distance-based price (e.g. xx CNY/km for the milage exceeds xx km). For trips originating from different areas, different distance-based price can be set for those areas, in order to balance the differentiated area-wise gap between supply and demand. Accordingly, the MDP for pricing can be adjusted as follows.

\textbf{Agent}: The ride-sourcing platform which makes the overall pricing decision.

\textbf{State}: For simplicity, grid system on the road network can be also adopted to represent different areas. The corresponding state for the platform is the collection of the number of available drivers, waiting passengers for each grid, together with the grid-wise distance-based prices. 

\textbf{Action}: The action for the platform is to set new distance-based prices for trips originating from different grids.

\textbf{Reward}: The reward can be set to the overall order revenue for the current time interval.

\textbf{State transition}: The transition possibility from the current platform state to the next one after actions made.

Although there are a variety of RL algorithms for different MDP settings, but their requirement for the simulators are basically the same. The simulator serves as an external environment to provide state transitions ($S$, $A$, $S^{'}$, $R$), for the repeating updation of RL models. To this end, the simulator should be 1) very accurate and close to the real-world scenario for the correct state transition records, and 2) flexible to extract the required state transitions to the RL algorithm module. For accuracy, the proposed simulator has been designed to mimic the real-world ride-sourcing market, ranging from the implementation on road network to the consideration of passengers/drivers behaviors. For flexibility, in the previous examples, the states of drivers or platform can be easily obtained via single agent properties or the collection of them, as shown in Fig. \ref{simulation framework}, while the related actions and rewards can be extracted in the matching, pricing and repositioning module. The abundance of the properties setting and the clarity of operation modules in the proposed simulator can assist in the fast generation of transition records for the training of RL algorithms for ride-sourcing operations.

\section{Experiments for theoretic model related tasks}

To validate the capability of the proposed simulator on theoretic model related task, we design this experiment of matching model comparison. In the experiment, we first set up some metrics, and compare their values under different theoretic matching functions and the simulation. Matching function depicts the relationship between the successful matching rate and the supply/demand related parameters, which is frequently utilized in ride-sourcing related studies. The closeness of the metrics deducted via the matching functions to the simulation is utilized to measure the accuracy of the matching function. Via the comparison, we can discover the proper matching function to use under different market status. Similar idea can also be extended to other ride-sourcing related studies, based on the proposed simulator. 

For experimental details, we set the simulation time interval to 5 seconds. The orders are generated from an order distribution pattern of Manhattan, including the passenger’s origin, destination, and order generation time. The vehicle speed is 6.33 $m/s$ and Passengers’ maximum waiting time is 300 $s$. We also set 2000 $m$ as maximum pick-up distance. Due to the small number of orders in the morning and early morning, we selected orders from 10:00 a.m. to 10:00 p.m. as experimental data to enhance the stability of the simulation. Arrival rate of orders per second is extracted and adjusted based on the real data. The fleet sizes range from 200 to 2000 with a step of 200 represent numbers of drivers. 

We combine 10 different arrival rates of new orders and 10 fleet sizes (200 to 1000) to simulate 100 sets of market scenarios. For each scenario, we implement our simulator and theoretical models to compare their matching process under different market situations. For comparison, we set up 6 theoretic matching function, including perfect matching, FCFS, Cobb-Douglas production function, M/M/1 (or N) queueing model, M/M/1/k queueing model, and batch matching model (An overall introduction on these models can be found in \cite{wei2022calibration}). 

For a full comparison, we employ four metrics to evaluate the simulation results of the simulator, as shown below. The closeness between simulation result and the theoretic result is computed via Mean Absolute Percentage Error (MAPE).

\begin{itemize}
\item \textbf{Matching rate}: Total matched orders divided by total produced orders.
\item \textbf{Average waiting time}: Passengers average waiting time before getting pick-up, consisting of matching time and pick-up time.
\item \textbf{Matching time}: Time between production and match of the orders.
\item \textbf{Pickup time}: Average effective orders’ pick-up time, which drivers spend on the way to pick up passengers.
\end{itemize}

\subsection{Matching model comparison}

\begin{figure}[t!]
	\centering
	\includegraphics[width=0.8\textwidth]{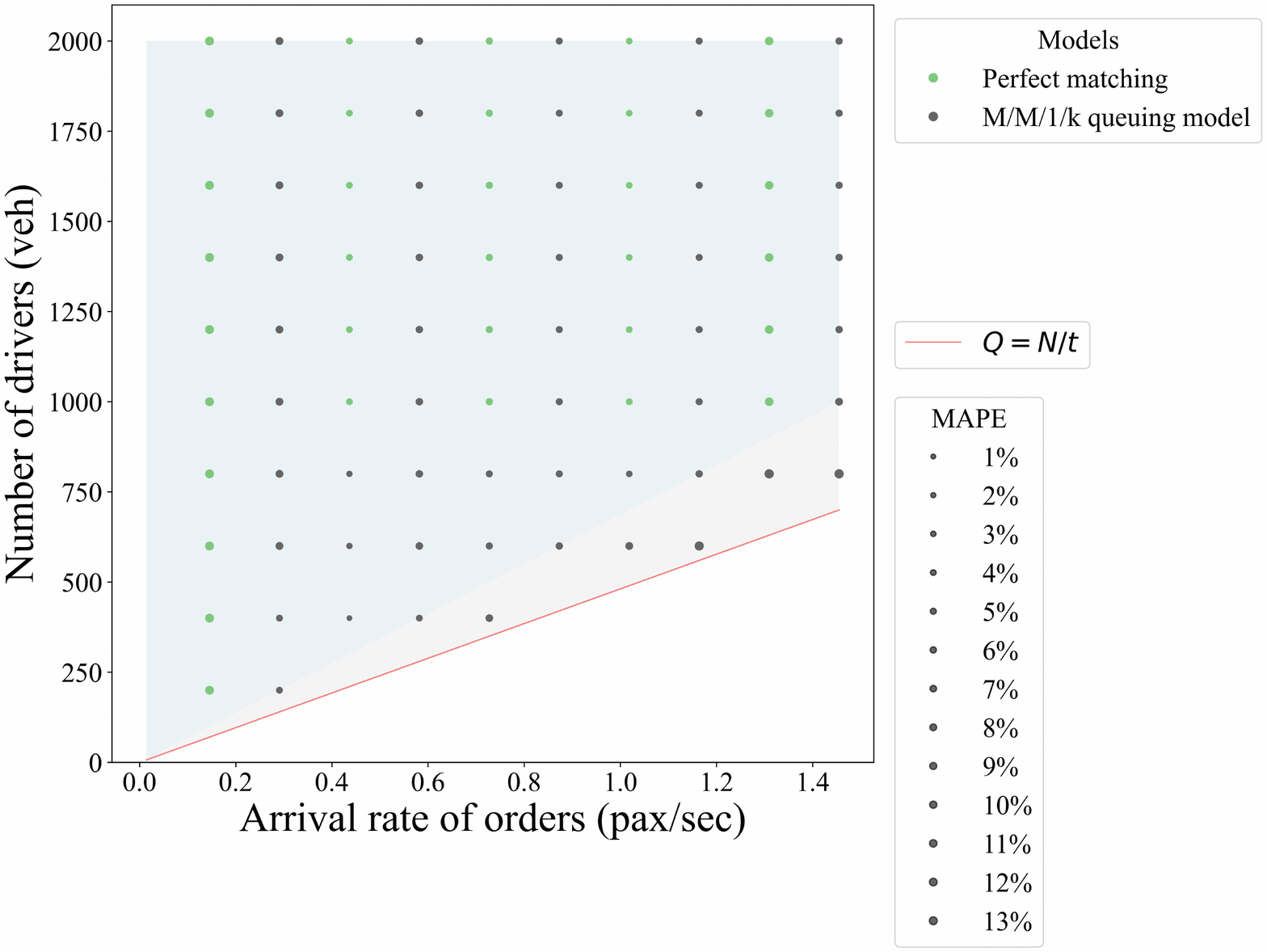}
	\caption{The best-fit models for matching rate estimation in different markets}
	\label{matching_rate_best_model}
\end{figure}

In order to evaluate which model is more suitable for calculating the matching rate in different market scenarios, we first compute MAPE of the matching rates generated by matching function and the simulation . The best-fit models for matching rate is shown in Fig. \ref{matching_rate_best_model}.

In Fig. \ref{matching_rate_best_model}, the horizontal axis stands for the arrival rate of passengers (demand), which varies from 0.01 to 1.455. The vertical axis stands for the number of drivers (supply), which vary from 200 to 1000.  The markets closer to the upper left corner have sufficient supply, and closer to the lower right corner have insufficient supply. Each point on the chart represents an equilibrium market under a combination of supply(on the y-axis) and demand(on the x-axis). The point in the figure represents the best-fit model for estimating the matching rate in a particular market, and the size of point means the MAPE value. Larger the size, larger error of MAPE value.

\begin{figure}
\begin{subfigure}{.5\textwidth}
\centering
\includegraphics[width=.9\linewidth]{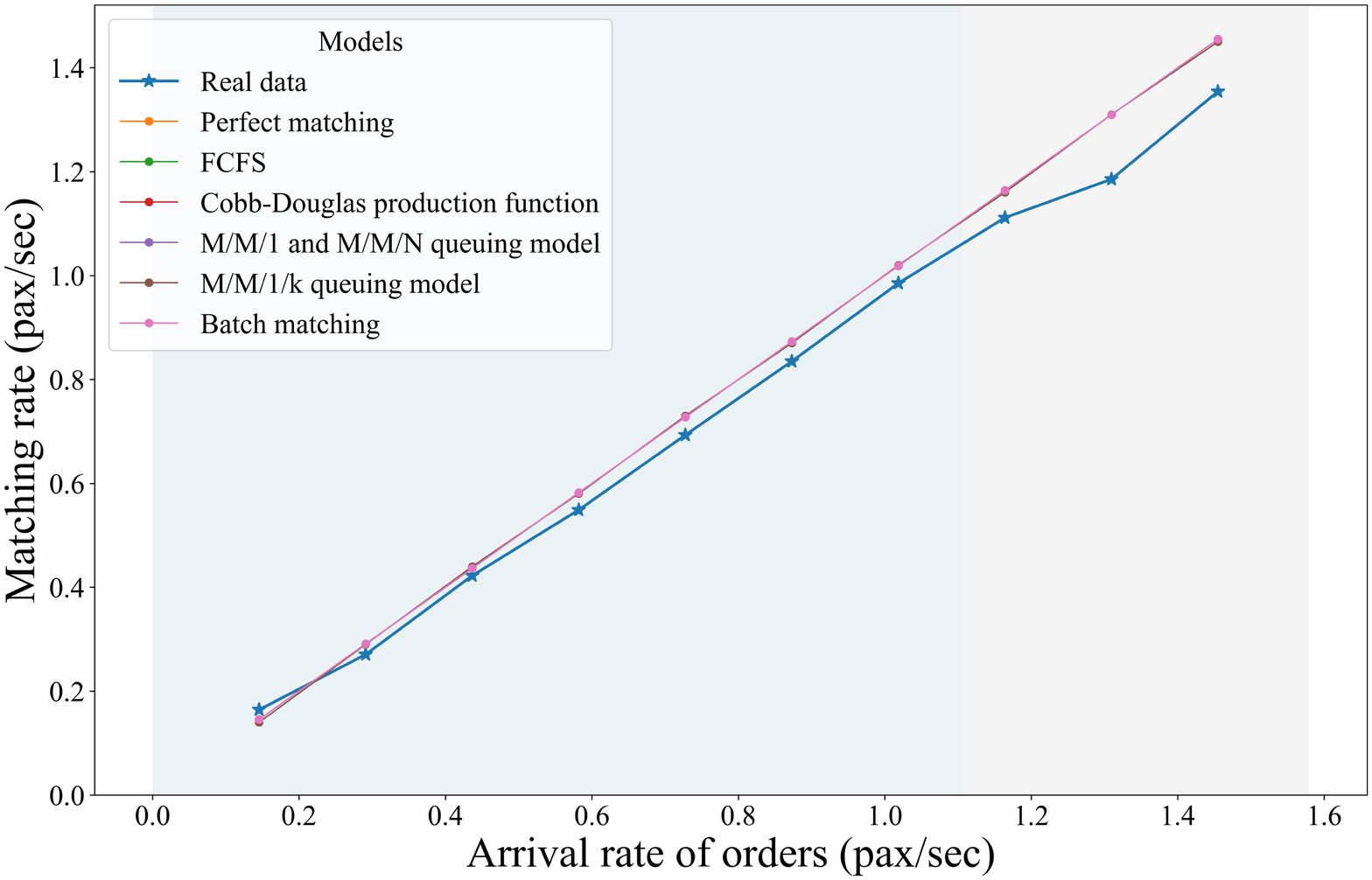}
\caption{Relationship between matching rate and demand under 1000 vehicles}
\label{Matching rate_fix_driver=1000}
\end{subfigure}%
\hspace{.3in}
\begin{subfigure}{.5\textwidth}
\centering
\includegraphics[width=.9\linewidth]{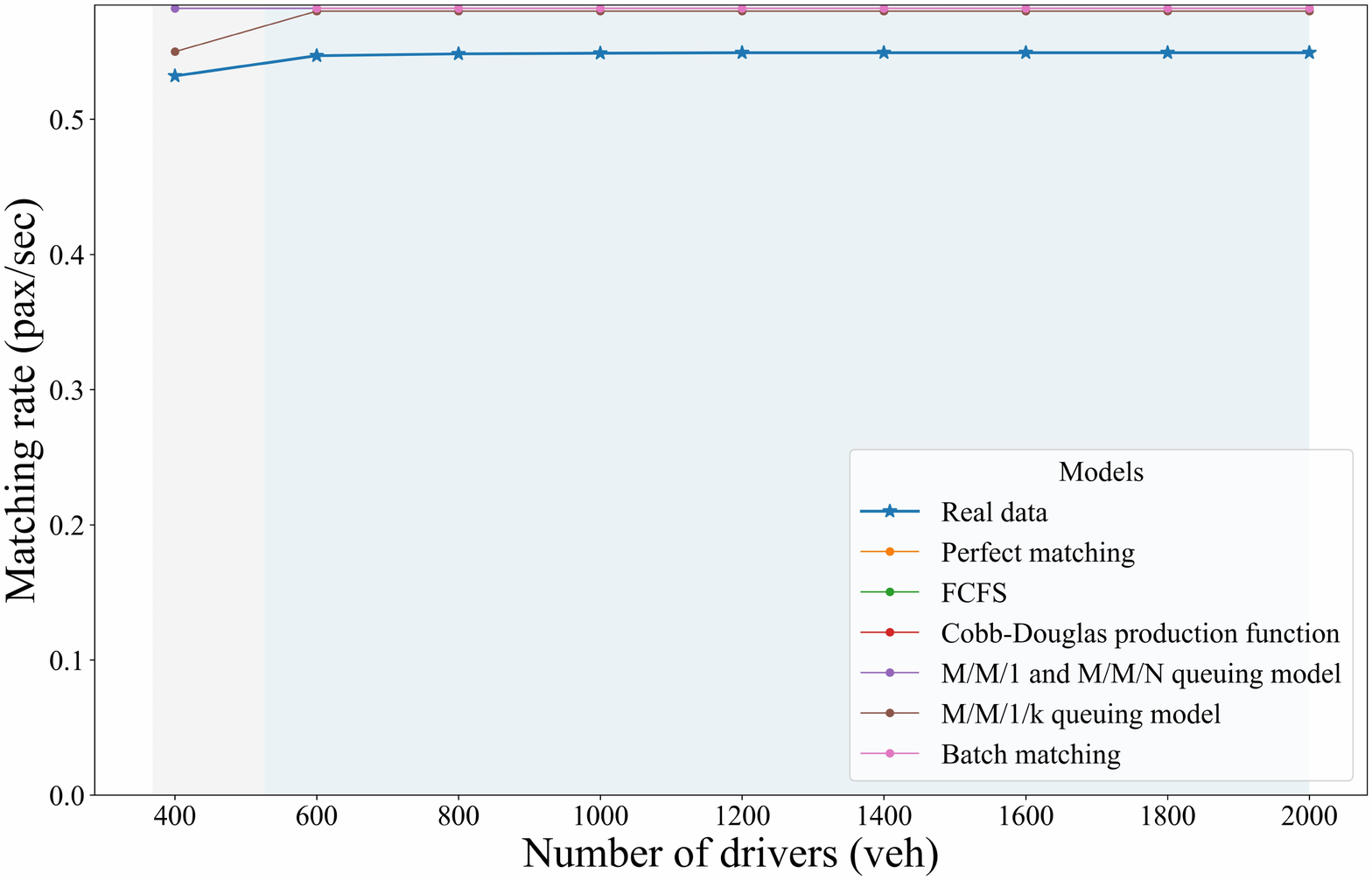}
\caption{Relationship between matching rate and supply under 0.58 pax/sec arrival rate of orders}
\label{Matching rate_fix_order=0.4}
\end{subfigure}\\[1ex]
\caption{Matching rate estimation under different supply-demand relationships}
\label{relation 1}
\end{figure}

Fig. \ref{matching_rate_best_model} shows that the perfect matching model and M/M/1/k Queuing Model simulate the real data better in most markets scenarios. We also fix either the fleet size or the arrival rate of orders, and draw the relationship between matching rate with another one. The result is shown in Fig. \ref{relation 1}, where the meaning of background color is consistent with above (blue for over-supplied markets and grey for balanced markets), the true values generated by our simulator are presented by the blue star symbol, and the estimated values are represented by the dot symbol. From Fig. \ref{relation 1}, we can easily find the trend of the true value of the matching rate and the difference between various models.

We can see that with the fleet size fixed at 1000 (Fig. \ref{Matching rate_fix_driver=1000}), as the arrival rate of orders increases, the true matching rate strictly increases with the input of the passenger arrival rate Q, but the growth rate first keeps stable then becomes slow. When the supply is sufficient (blue area), all requests will be served, and the matching rate equals the arrival rate of orders. When the demand continues to increase, the matching rate is constrained by insufficient supply. Except M/M/1/k queuing model, other models all implicitly assumed that every arrival passenger gets served finally and estimate the matching rate by Q, which is a given input in each market scenario. 

In over- supplied markets, all models can estimate the matching rate with a small error because almost all passengers can get served as supply is sufficient. However, when the supply becomes insufficient and only part of the passengers are served, the assumption raised before may lead to larger errors. To summarize, the Cobb-Douglas production functions, M/M/1 and M/M/N queuing models, FCFS, batch matching models are more proper for oversupplied markets and partially equilibrium markets. Perfect matching and M/M/1/k queuing models are appropriate in a wider scope of supply situation, with respect to matching rate for ride-sourcing market. 

\subsection{Best-fit models for matching time estimation}

\begin{figure}[t!]
	\centering
	\includegraphics[width=0.8\textwidth]{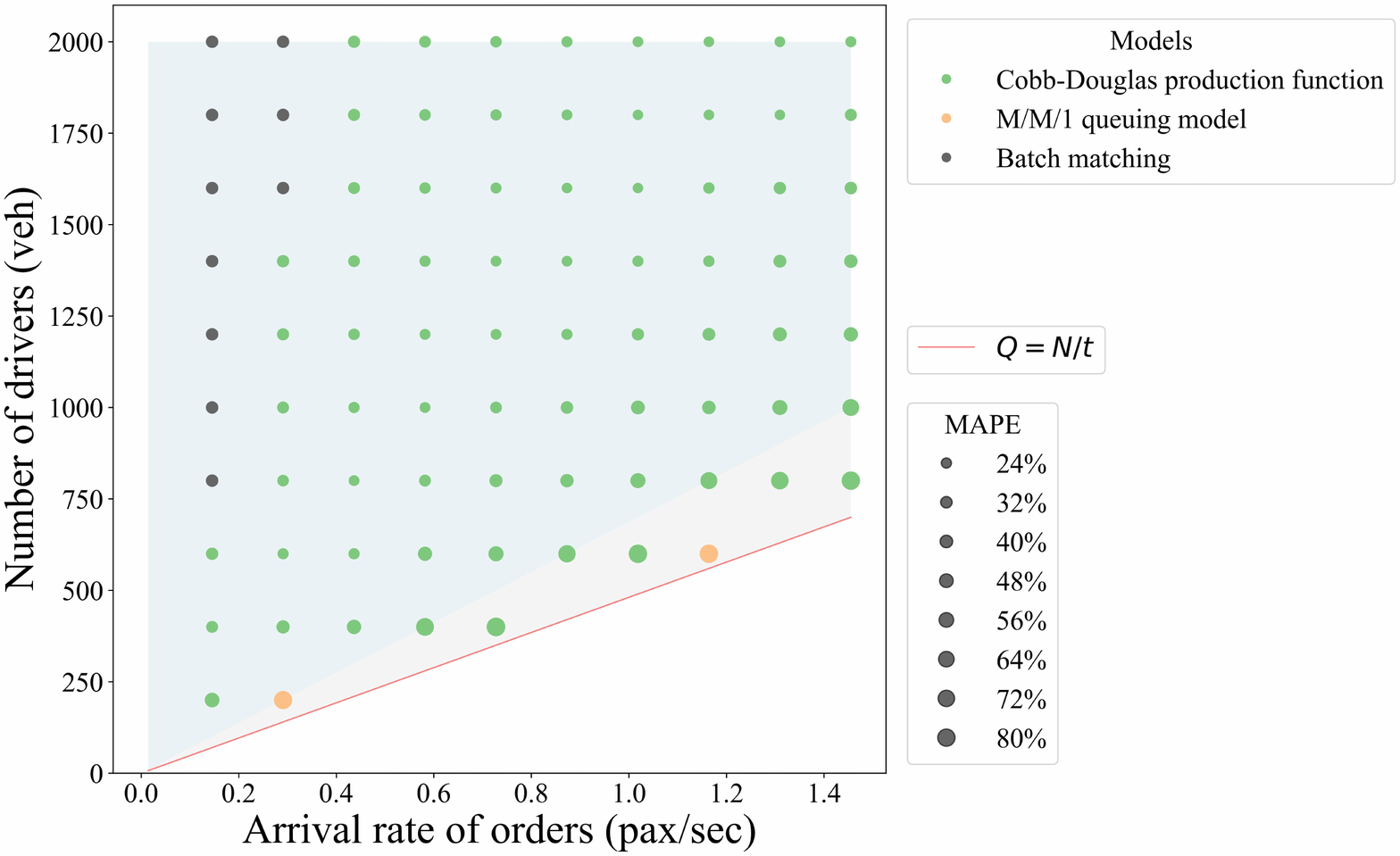}
	\caption{The best-fit models for matching time estimation in different markets}
	\label{matching_time_best_model}
\end{figure}

Fig. \ref{matching_time_best_model} shows the best-fit models for matching time estimation in each market. Overall, Cobb- Douglas production function is suitable for the balanced markets and over- supplied markets. In the over-supplied markets (blue area), the matching time of most theoretical models is close to the simulation result. However, in the balanced markets (grey area), the real matching time become larger and  more hard to estimate. Observed from the figure, Cobb- Douglas production function is very close to the true value in terms of both trend and absolute value. However, this model becomes unsolvable when the supply in the market starts to be inadequate. When the market is close to the red line, namely, Q approaching N=t, which means passenger arrival rate Q is equal to the vacant vehicle arrival rate (estimated by service capacity N=t), most models become infeasible. In such scenario, M/M/1/k queuing model, which considers the passenger departures, is more consistent with the market.

\begin{figure}
\begin{subfigure}{.5\textwidth}
\centering
\includegraphics[width=0.9\linewidth]{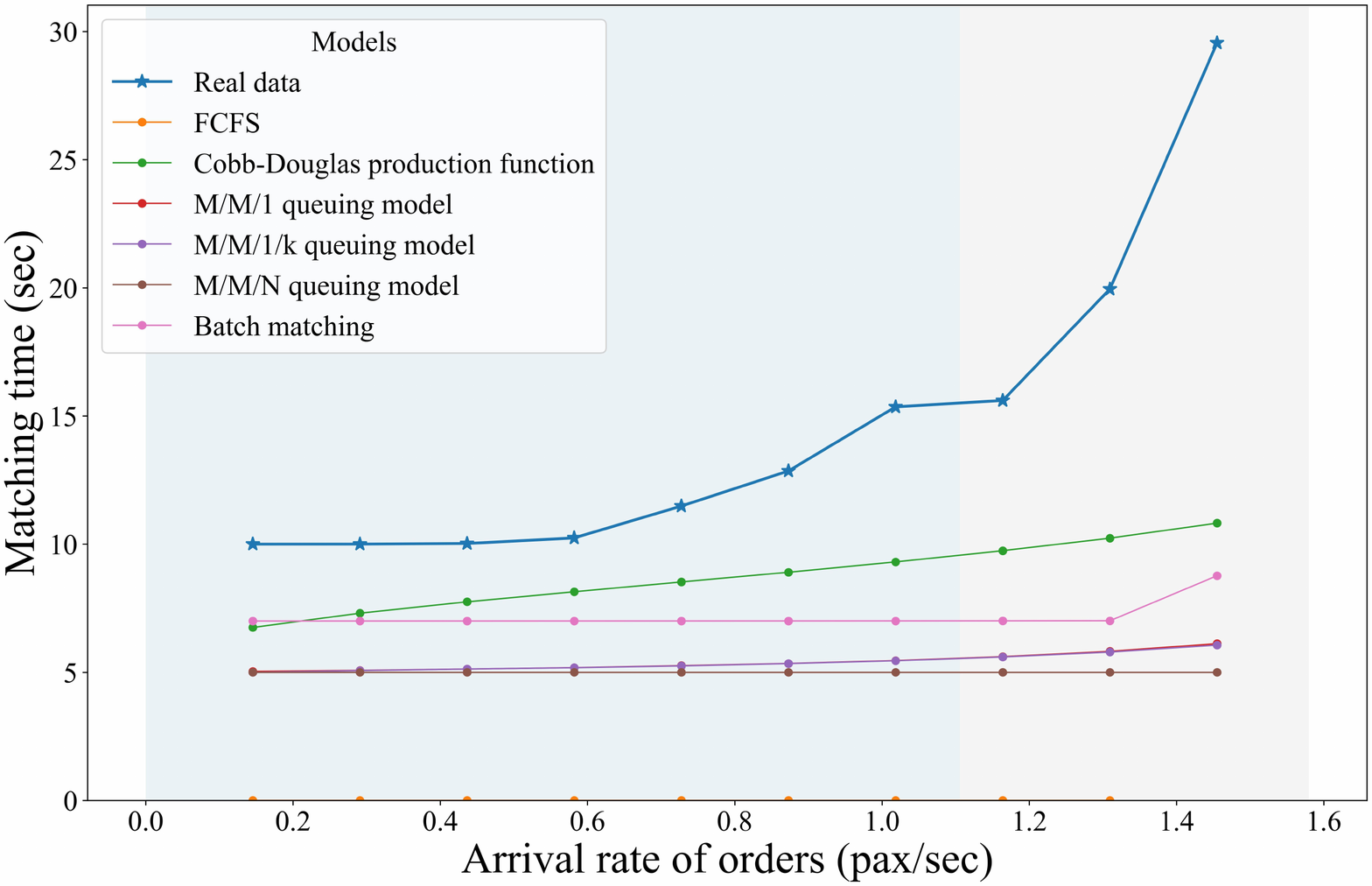}
\caption{Relationship between matching time and demand under 1000 vehicles}
\label{Matching time_fix_driver=1000}
\end{subfigure}%
\hspace{.3in}
\begin{subfigure}{.5\textwidth}
\centering
\includegraphics[width=0.9\linewidth]{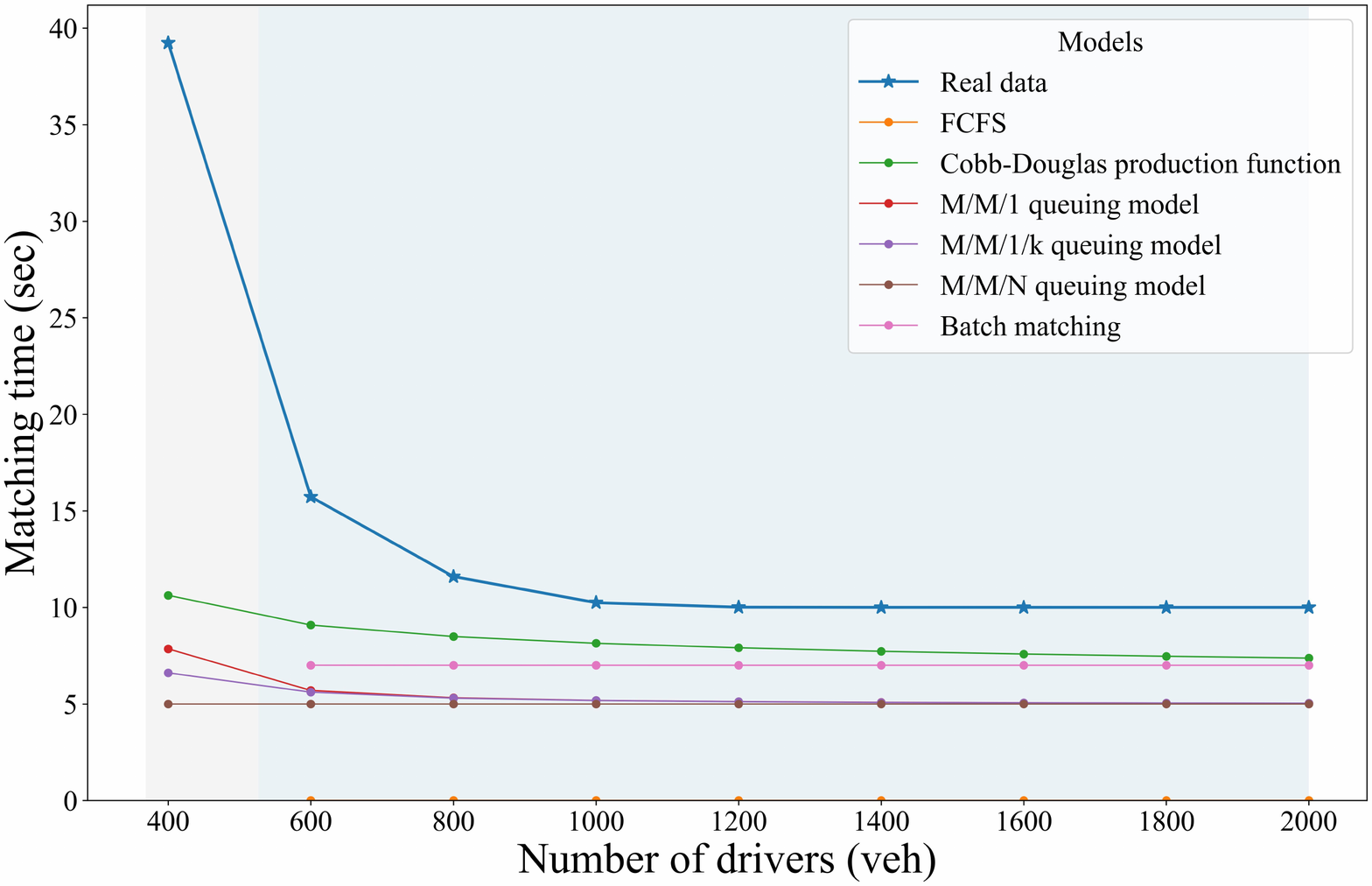}
\caption{Relationship between matching time and supply under 0.58 pax/sec arrival rate of orders}
\label{Matching time_fix_order=0.4}
\end{subfigure}\\[1ex]
\caption{Matching time estimation under different supply-demand relationships}
\label{relation 2}
\end{figure}

\subsection{Best-fit models for pick-up time estimation}

\begin{figure}[t!]
	\centering	\includegraphics[width=0.8\textwidth]{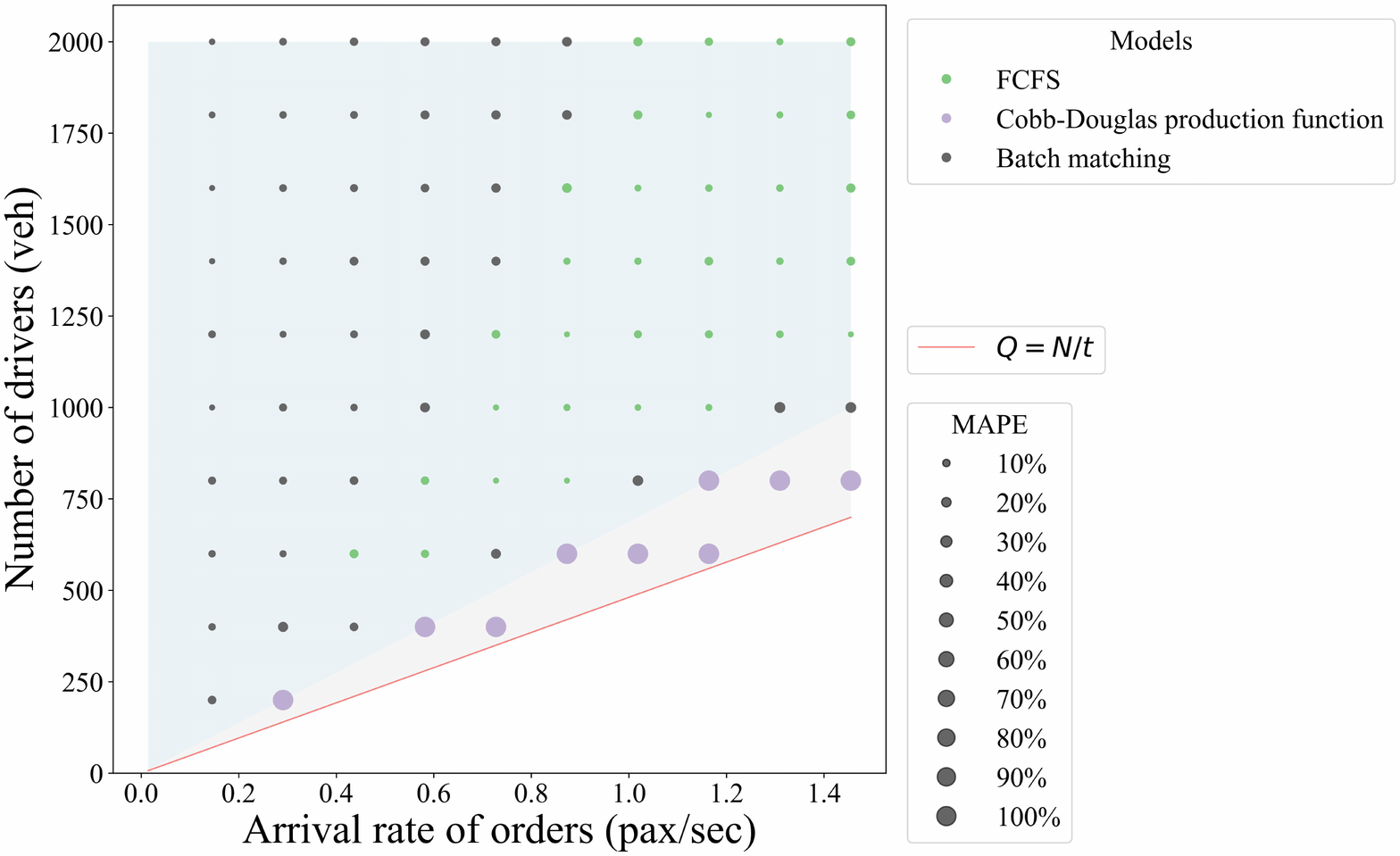}
	\caption{The best-fit models for pick-up time estimation in different markets}
	\label{pickup_time_best_model}
\end{figure}

Fig. \ref{pickup_time_best_model} shows the best-fit models for pick-up time estimation in different markets. Overall, FCFS model and Batch Matching are suitable for the over-supplied markets, and Cobb-Douglas production function is more proper for the balanced markets.  However, in the balanced market, the MAPE between Cobb-Douglas production function and our simulator is still huge. Possible reasons are that a fixed max pickup distance is set. Only when the pick-up distance is less than it, the simulator will dispatch some driver to the order. To this end, the pick-up time has a maximum actually in the simulator, while the theoretical model do not consider this point. Still, FCFS model and batch matching model are more suitable to calculate the pick-up time since they well characterize drivers’ idle, pick-up, and in-trip phases. Moreover, batch matching model considers both pick- up time and matching time, while FCFS focuses on characterizing the pick-up time and regards the matching time as zero. 

\begin{figure}
\begin{subfigure}{.5\textwidth}
\centering
\includegraphics[width=0.9\linewidth]{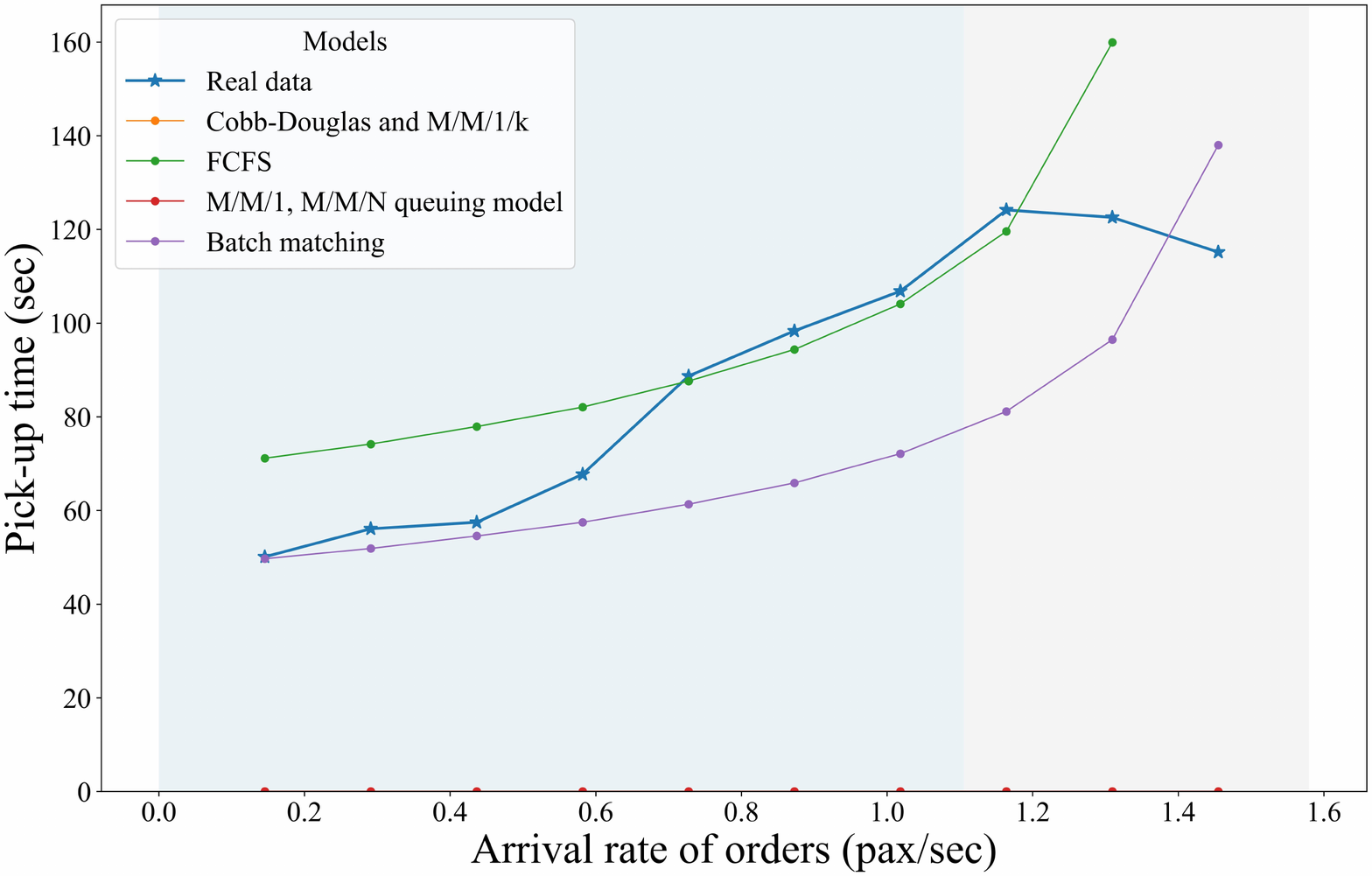}
\caption{Relationship between pick-up time and demand under 1000 vehicles}
\label{Pick-up time_fix_driver=1000}
\end{subfigure}%
\hspace{.3in}
\begin{subfigure}{.5\textwidth}
\centering
\includegraphics[width=0.9\linewidth]{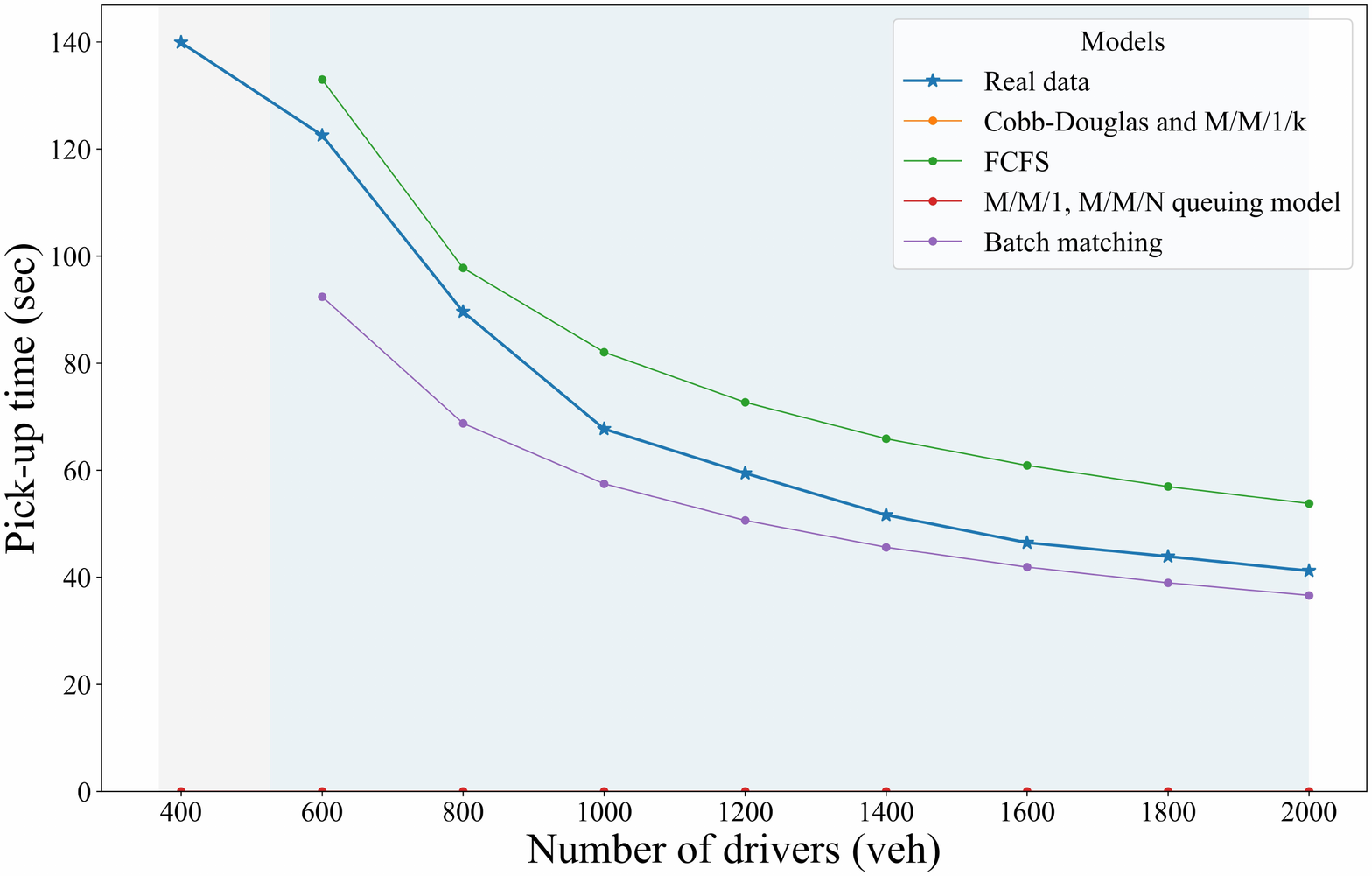}
\caption{Relationship between pick-up time and supply under 0.58 pax/sec arrival rate of orders}
\label{Pick-up time_fix_order=0.4}
\end{subfigure}\\[1ex]
\caption{Pick-up time estimation under different supply-demand relationships}
\label{relation 3}
\end{figure}

\section{Experiments for reinforcement learning based tasks}

\subsection{Data and experiment design}

In this part, we aim to show the capability of the proposed simulator to implement different types of reinforcement learning (RL) tasks for ride-sourcing service. We mainly focus on the matching and idle vehicle repositioning RL tasks, respectively. For basic setting, the utilized data is extracted from the trip record of NYC TLC Yellow Taxi dataset for May, 2015. The studied area is a square containing Manhattan in the center, where the grid system is also constructed as squares. The dataset contains trip-specific information, including origin, destination, starting time of trips, etc. The studied time period ranges from 3 a.m. to 9 a.m., and the length of each time interval for matching in simulation is 1 min. Training set for RL algorithms include dates from May 4, 2015 to May 8, 2015, while the testing set include dates from May 11, 2015 to May 15, 2015. The speed of vehicle is 6.33 $m/s$. The maximum pick-up distance is 1000 m, and the matching mode is bipartite matching. Behaviors of passengers and drivers are also considered, where the maximum waiting time for passengers is five minutes, beyond which passengers will cancel their orders. For simplicity, we bootstrap 10 percent of the original orders for simulation, and set 100 drivers in the market. Fare of trips is obtained based on the taxi trip fare structure by Taxi \& Limousine Commission in NYC, where the optimization goals for both experiments are the total revenue maximization for the ride-sourcing platform.

In addition to the general experimental setting above, we also make some specific design for the two experiments respectively. For matching related experiment, the definition of MDP is similar to that in section 4.2. To solve the MDP, we adopt a value table based reinforcement learning algorithm, where we specify the value function via repetitive data-driven learning, and utilize it to capture the future gain of different matching decision. More details can be found in \cite{xu2018large}, whose logic we mainly follow. In addition, we also design several different baselines to compare their performances, including Myopic, and Pick-up distance based (PDB). The first baseline optimize instant order revenue for a certain time interval. The PDB is similar to the Myopic method, except that its goal becomes minimizing the overall pick-up distance. Both of the baselines do not consider the future gain when making current decisions, while the RL method involves the future gain. 

For repositioning related experiment, the basic setting is the same as in matching related experiment, except that we focus on the optimization of idle vehicle repositioning decision. As discussed in section 4.2, the agents of the repositioning MDP are the drivers, and the action space is for the idle drivers to move to the neighboring grids around their current locations. Specially, the reward is set to the average order revenue of the destination grid for repositioning at the time when the driver is expected to arrive. In this way, excessive accumulation of idle drivers on some hot area can be avoided. To solve this MDP, we adopt an regular RL algorithm called Actor-Critic (A2C), which was first developed by \cite{lin2018efficient} for vehicle repositioning. To compare the performances of RL and other approaches for repositioning task, a baseline is designed where the idle vehicle randomly move to surrounding areas without special guidance.

For more straightforward comparison within different methods, we employ several performance metrics for both experiments, as listed below.

\begin{itemize}
\item \textbf{Platform revenue}: Average of total platform revenue over the testing days, with US dollar as unit.
\item \textbf{FRAO}: Fulfillment rate of all orders.
\item \textbf{Occupancy rate}: Average occupancy rate of vehicles. 
\item \textbf{Matching time}: Average waiting time of passengers before getting matched, with second as unit.
\item \textbf{Pickup time}: Average pickup time for matched passengers and drivers, with second as unit.
\end{itemize}

\subsection{Results and analysis}

The results for matching RL experiment are shown in Table 2. The tested RL method outperforms the best of the other two baselines in both platform revenue and occupancy rate by 10.8\% and 6.4\%, respectively. Although the matching time and pickup time metrics for RL method is not the best, the excessive part of these two metrics is still acceptable for passengers. In addition, the results for repositioning RL experiment are shown in Table 3. The tested RL method is superior than the baseline for all the metrics. Specially, it outperforms the baseline in platform revenue and occupancy rate by 12.9\% and 6.1\%, demonstrating its capability to guidance the vehicle movement more effectively. In summary, we can train, test and analyze the RL algorithms based on the proposed simulator in an efficient way. It is worth mentioning that other RL approaches can be also fast trained and tested on our open-sourced simulation platform for different operations of ride-sourcing services.

\begin{table}
\begin{center}
\caption{Comparison between RL methods and baselines for matching operation}
\begin{tabular}{c|c|c|c|c|c}
    \toprule[2pt]
    \textbf{Method} & \textbf{Platform revenue} & \textbf{FRAO} & \textbf{Occupancy rate} &\textbf{Matching time} &\textbf{Pickup time} \\
    \toprule[1.2pt]   
    Myopic & 10978.8 & 10.0\% &46.8\% & 155.2 & 176.6\\
    \hline
    PDB & 11635.8 & 17.9\% &47.0\% & 179.8 & 109.5 \\
    \hline
    \textbf{RL} &\textbf{12889.1} &  \textbf{17.9\%} &  \textbf{53.2\%} &  \textbf{189.8} &  \textbf{153.1}\\
    \toprule[2pt]
\end{tabular}
\end{center}
\label{matching rl}
\end{table}

\begin{table}
\begin{center}
\caption{Comparison between RL methods and baselines for idle vehicle repositioning}
\begin{tabular}{c|c|c|c|c|c}
    \toprule[2pt]
    \textbf{Method} & \textbf{Platform revenue} & \textbf{FRAO} & \textbf{Occupancy rate} &\textbf{Matching time} &\textbf{Pickup time} \\
    \toprule[1.2pt]   
    Random & 11523.5 & 10.5\% &49.5\% & 153.4 & 181.5\\
    \hline
    \textbf{RL} &\textbf{13007.0} &  \textbf{12.2\%} &  \textbf{55.6\%} &  \textbf{152.7} &  \textbf{172.1}\\
    \toprule[2pt]
\end{tabular}
\end{center}
\label{repo rl}
\end{table}

\section{Conclusion}
This paper develops a comprehensive, multi-functional and open-sourced simulation framework for ride-sourcing service operations. The simulation framework is constructed following the natural operation process of a ride-sourcing platform, and can be run on the actual road network rather than artificial grid systems, with a routing module for calculating the shortest paths. Drivers and passengers' heterogeneous behaviors, such as passengers' sensitivity to price and waiting times, and drivers' working schedules over a day, are well considered and integrated to the simulation platform. The simulator is also designed for fast modification and implementation of different tasks, including the validations of theoretical models, trainings of reinforcement learning models for matching and repositioning. A visualization module with different modes is also developed for the demonstration of different demand and supply status in the market. The capability of the simulator is validated via three representative tasks, including evaluating theoretical models in the literature and training RL models for on-demand matching and repositioning.  

In the future, the simulation framework can be further improved via the following perspectives. First, the simulator can be extended with transit network, autonomous vehicle agents, electric vehicle agents, and congestion computation module to mimic the multi modal transportation market. Second, the simulation can be modified to accommodate the ride-pooling service. Third, more advanced algorithms for ride-sourcing service operations can be developed and tested in the proposed simulation platform.

\bibliographystyle{apalike}
\bibliography{main}
\end{document}